\documentclass[10pt,twocolumn,letterpaper]{article}
\usepackage{titletoc} % Alternative package

\usepackage{tocloft} % Load the tocloft package without options

\usepackage[pagenumbers]{cvpr} % To force page numbers, e.g. for an arXiv version

\usepackage{multirow}
\usepackage{graphicx}
\usepackage{times}
\usepackage{epsfig}
\usepackage{amsmath}
\usepackage{amssymb}
\usepackage{xcolor}
\usepackage{comment}
\usepackage{bbm}

\usepackage[hypcap=false]{caption}

\definecolor{cvprblue}{rgb}{0.21,0.49,0.74}
\usepackage[pagebackref,breaklinks,colorlinks,allcolors=cvprblue]{hyperref}

\def\paperID{522} % *** Enter the Paper ID here
\def\confName{CVPR}
\def\confYear{2025}

\newcommand{\JP}[1]{{\color{blue}  [{JP: #1}]}}
\newcommand\pl[1]{{\color{violet}#1}}

\title{OVGaussian: Generalizable 3D Gaussian Segmentation with Open Vocabularies}
\author{Runnan Chen$^{1}$~\quad Xiangyu Sun$^{1}$~\quad Zhaoqing Wang$^{1}$\\ Youquan Liu$^{2,7}$  ~\quad Jiepeng Wang$^{3}$~\quad Lingdong Kong$^{4,7}$ ~\quad Jiankang Deng$^{5}$\\ Mingming Gong$^{6}$ ~\quad Liang Pan$^{7}$ ~\quad Wenping Wang$^{8}$ ~\quad Tongliang Liu$^{1}$
\\[0.2ex]
\small{$^{1}$The University of Sydney} \quad 
\small{$^{2}$
Fudan University} \quad 
\small{$^{3}$The University of Hong Kong} \quad 
\small{$^{4}$National University of Singapore}\\
\small{$^{5}$Imperial College London} \quad
\small{$^{6}$The University of Melbourne} \quad
\small{$^{7}$Shanghai AI Laboratory} \quad
\small{$^{8}$Texas A\&M University}
}

\begin{document}
% \maketitle

\twocolumn[{
    \renewcommand\twocolumn[1][]{#1}
    \maketitle
    \centering
    \vspace{-0.6cm}
    \includegraphics[width=0.93\textwidth]{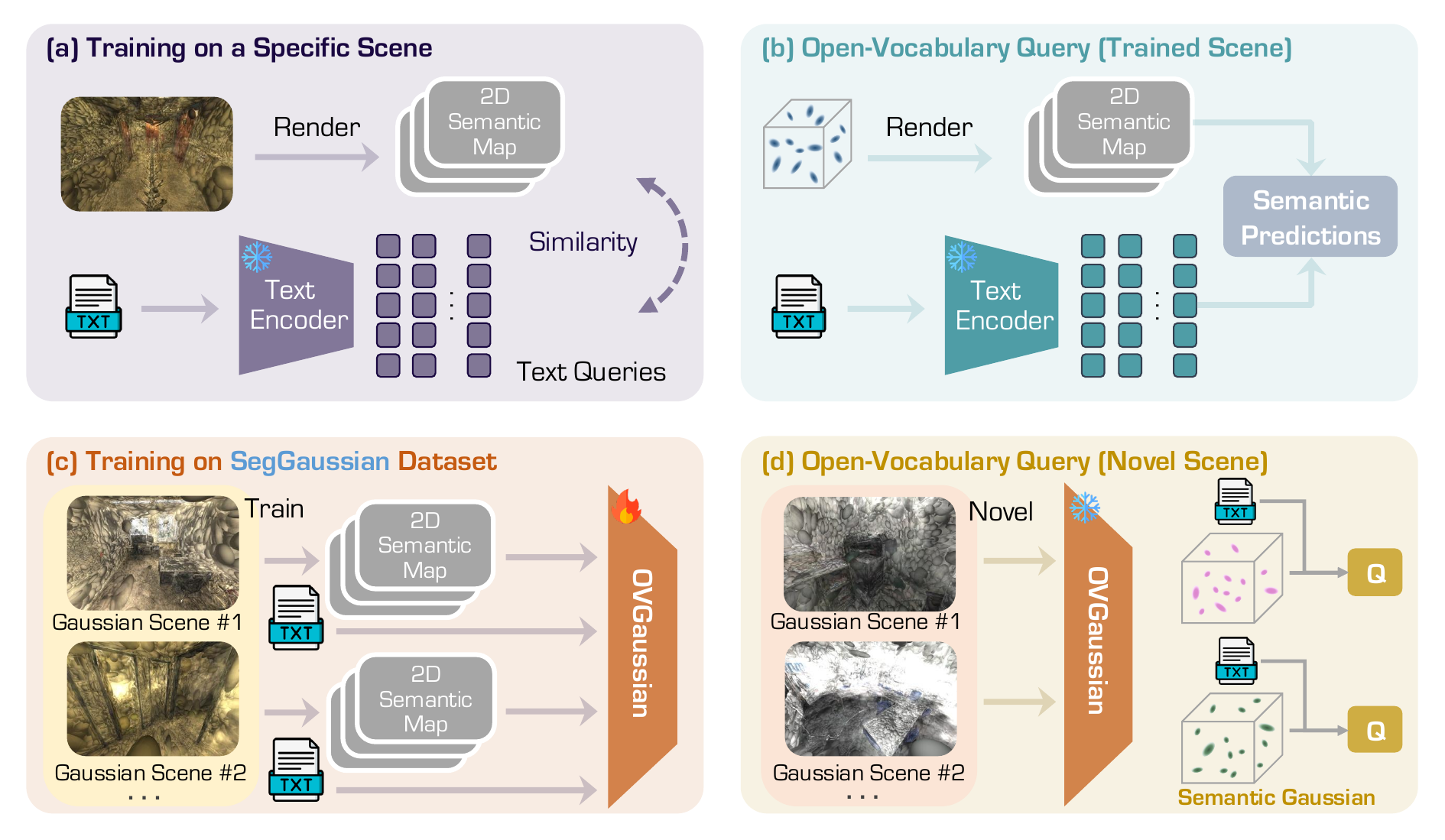}
    \vspace{-0.3cm}
    \captionof{figure}{We introduce \textbf{OVGaussian}, a novel approach that extends Gaussian-based representations for open-vocabulary semantic generalization across scenes. Unlike previous methods (upper part: \textbf{(a)} $\rightarrow$ \textbf{(b)}) that restrict open-vocabulary querying to specific trained scenes, OVGaussian (lower part: \textbf{(c)} $\rightarrow$ \textbf{(d)}) is trained on the \textbf{SegGaussian} dataset, enabling it to directly predict semantic property for each Gaussian in novel scenes, thereby achieving cross-scene open-vocabulary query.}
    \label{fig:teaser}
    \vspace{0.35cm}
}]

\begin{abstract}
Open-vocabulary scene understanding using 3D Gaussian (3DGS) representations has garnered considerable attention. However, existing methods mostly lift knowledge from large 2D vision models into 3DGS on a scene-by-scene basis, restricting the capabilities of open-vocabulary querying within their training scenes so that lacking the generalizability to novel scenes. In this work, we propose \textbf{OVGaussian}, a generalizable \textbf{O}pen-\textbf{V}ocabulary 3D semantic segmentation framework based on the 3D \textbf{Gaussian} representation. We first construct a large-scale 3D scene dataset based on 3DGS, dubbed \textbf{SegGaussian}, which provides detailed semantic and instance annotations for both Gaussian points and multi-view images. To promote semantic generalization across scenes, we introduce Generalizable Semantic Rasterization (GSR), which leverages a 3D neural network to learn and predict the semantic property for each 3D Gaussian point, where the semantic property can be rendered as multi-view consistent 2D semantic maps. In the next, we propose a Cross-modal Consistency Learning (CCL) framework that utilizes open-vocabulary annotations of 2D images and 3D Gaussians within SegGaussian to train the 3D neural network capable of open-vocabulary semantic segmentation across Gaussian-based 3D scenes. Experimental results demonstrate that OVGaussian significantly outperforms baseline methods, exhibiting robust cross-scene, cross-domain, and novel-view generalization capabilities. Code and the SegGaussian dataset will be released. \footnote{\url{https://github.com/runnanchen/OVGaussian}.}

\end{abstract}

\section{Introduction}
\label{sec:introduction}
% \begin{figure*}
%   %%%%\vspace*{-5ex}
%   \centerline{\includegraphics[width=0.85\textwidth]{v0/figures/teaser.png}}
%   %\vspace*{-0.5ex}
%   \vspace{-1.5ex}
%   \caption{We introduce OVGaussian, a novel approach that extends Gaussian-based representations for open-vocabulary semantic generalization across scenes. Unlike previous methods (upper part) that restrict open-vocabulary querying to specific trained scenes, OVGaussian (lower part) is trained on the SegGaussian dataset, enabling it to directly predict semantic property for each Gaussian in novel scenes, thereby achieving cross-scene semantic generalization.}
%   \vspace{-2ex}
% \end{figure*}

Open-vocabulary scene understanding has emerged as a crucial capability in computer vision, enabling models to recognize a wide variety of semantic categories, even those unseen during training \cite{chen2023clip2scene,peng2023openscene}. Recent efforts have extended open-vocabulary capabilities to 3D data, particularly 3D Gaussian representations \cite{kerbl20233d}, which efficiently capture the spatial and semantic properties of complex scenes. 

Current approaches to open-vocabulary 3D scene understanding using 3D Gaussians often adopt a ``lift-and-adapt" strategy, extracting features from large 2D models (Fig. \ref{fig:teaser}), such as CLIP \cite{radford2021learning}, and projecting them onto 3D representations to preserve the semantic richness learned from 2D images \cite{shi2024language,zhou2024feature,qin2024langsplat,ye2025gaussian}. While these approaches have shown some success, they face several critical limitations. Firstly, these methods are generally constrained to the specific scenes on which they were trained, limiting their adaptability to generalize across unseen 3D scenes. Besides, as 2D projections cannot fully provide 3D spatial relationships, transferring knowledge from 2D to 3D fails to capture the full geometric context necessary for accurate 3D spatial understanding. Another limitation is the lack of a unified framework for effectively integrating multimodal data in a way that maintains semantic consistency across both 2D and 3D domains, resulting in inconsistencies that degrade the quality of open-vocabulary segmentation.

To address these challenges, we propose OVGaussian, a novel approach designed to empower 3D Gaussian representations with open-vocabulary segmentation capabilities that generalize across diverse scenes. To achieve this, we constructed SegGaussian, a comprehensive dataset containing 288 3D scenes represented as 3D Gaussians, each annotated with semantic and instance labels for both Gaussian points and multi-view images. The SegGaussian dataset provides a rich foundation for training models capable of understanding 3D scenes from multiple viewpoints and semantic contexts. 

Leveraging the complementary nature of 2D and 3D data alongside open-vocabulary semantic descriptors, OVGaussian builds a model that transcends scene-specific limitations, enabling seamless open-vocabulary segmentation across diverse Gaussian-based 3D scenes. To achieve semantic generalization for Gaussian representations across scenes, we introduce Generalizable Semantic Rasterization (GSR). This method uses 3D Gaussians as inputs to a 3D neural network that predicts semantic property for each 3D Gaussian. Similar to the colour property in 3D Gaussians, these semantic property can be rendered into 2D semantic maps from various viewpoints via alpha blending. Furthermore, to equip the 3D Gaussians with open-vocabulary segmentation capabilities, we propose Cross-modal Consistency Learning (CCL) to train this 3D neural network. We leverage the semantic annotations of Gaussians and multi-view images within the SegGaussian dataset to align the semantic property of 3D Gaussians and their rendered 2D semantic maps with corresponding text embeddings. Additionally, we utilize CLIP's visual encoder to align CLIP's visual embeddings of 2D images with the rendered 2D semantic maps. This cross-modal alignment facilitates a shared semantic understanding between the 3D Gaussian representations and text embeddings, thereby enhancing the model's generalization and open-vocabulary segmentation capabilities across various 3D scenes.

Experimental results demonstrate that OVGaussian achieves state-of-the-art performance in open-vocabulary segmentation, showcasing its effectiveness in cross-scene, cross-domain, and novel-view generalization. Our work establishes a promising new direction for open-vocabulary understanding in 3D spaces, making Gaussian-based representations versatile tools for semantic segmentation across diverse real-world scenarios.

The key contributions of our work are as follows.
\begin{itemize}

\item {We introduce SegGaussian, a dataset with 288 3D Gaussian scenes and comprehensive semantic annotations, providing a foundation for cross-scene 3D Gaussian understanding.}

\item {We propose Generalizable Semantic Rasterization (GSR), enabling 3D Gaussians to generalize across scenes by predicting semantic property that can be rendered into 2D semantic maps.}
\item {Cross-modal Consistency Learning (CCL) aligns 3D Gaussians with 2D maps and text embeddings, enhancing open-vocabulary segmentation across different scenes and viewpoints.}

\item {OVGaussian achieves state-of-the-art performance in open-vocabulary segmentation, demonstrating strong generalization across diverse scenes, domains, and novel views.}

\end{itemize}

\section{Related Work}
\label{sec:relatedwork}

\begin{figure*}
  %%%%\vspace*{-5ex}
  \centerline{\includegraphics[width=1\textwidth]{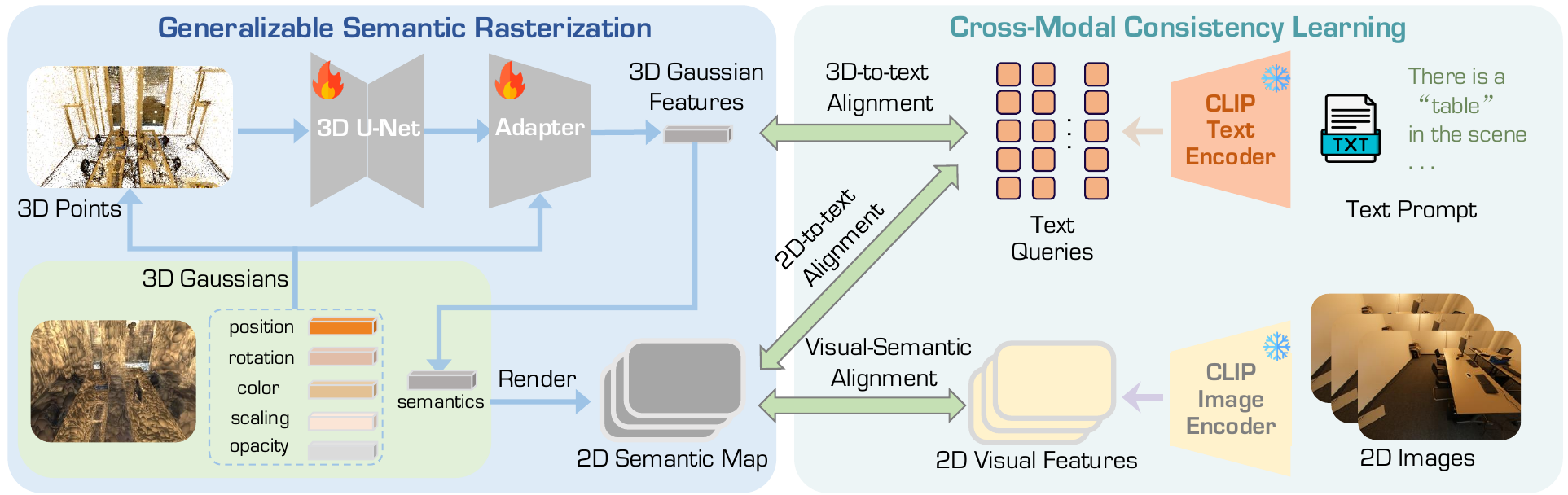}}
  %\vspace*{-0.5ex}
  \vspace{-1.5ex}
  \caption{Illustration of the \textbf{OVGaussian} framework. Our approach combines Generalizable Semantic Rasterization (GSR) to predict semantic properties for 3D Gaussians and Cross-modal Consistency Learning (CCL) to align these properties with vocabulary embeddings and 2D visual features. Trained on the SegGaussian dataset, OVGaussian enables cross-scene, open-vocabulary segmentation, achieving robust semantic generalization across diverse 3D scenes and viewpoints.}
  \label{fig:framework}
  \vspace{-2ex}
\end{figure*}

\paragraph{Scene Understanding.}
Scene understanding, focused on recognizing objects and spatial relationships, is central to applications in robotics, autonomous vehicles, and urban intelligence. Supervised methods have achieved notable results in 2D and 3D scene analysis~\cite{zhu2020cylindrical, wu2022point, cheng2020panoptic, qi2017pointnet, zhu2021cylindrical, rpvnet, af2s3net, kong2023rethinking, hong2022dsnet, cheng2021per, strudel2021segmenter, cheng2022masked,sun2024empirical,yin2024fusion,xu2023human,kong2023robo3d,contributors2020mmdetection3d,liu2023uniseg}, yet they rely heavily on large-scale, labor-intensive annotations, limiting their adaptability to new object categories. Open-world scene understanding methods~\cite{chen2023bridging,chen2022zero, michele2021generative, ding2022language, riz2023novel, xu2021simple, lilanguage, bucher2019zero, li2020consistent, hu2020uncertainty, zhang2021prototypical,liu2024multi,chen2023bridging,lu2023see,chen2023model2scene} seek to address this by identifying unseen categories without specific training data. Other works~\cite{peng2023openscene, XuYan20222DPASS2P, chen2023clip2scene, sautier2022image, chen2024towards, peng2025learning} reduce 3D annotation needs by leveraging 2D knowledge, expanding 3D segmentation with lower labeling costs. Vision foundation models like CLIP~\cite{radford2021learning} and SAM~\cite{kirillov2023segment} have further advanced open-world tasks, facilitating the transfer of rich 2D knowledge into 3D representations such as point clouds, neural fields, and 3D Gaussians for label-free 3D scene understanding. Recent models, including CLIP2Scene~\cite{chen2023clip2scene} and CNS~\cite{chen2024towards}, have improved 3D scene comprehension by incorporating 2D-3D calibration based on CLIP and SAM. While most prior work has focused on 3D point clouds, this study explores open-vocabulary segmentation on 3D Gaussians, emphasizing generalization across scenes, domains, and novel viewpoints.

\vspace{-3ex}
\paragraph{3D Gaussian Splatting.}
3D Gaussian Splatting~\cite{kerbl20233d,chen2024beyond} has emerged as a highly effective approach for real-time radiance field rendering to reconstruct 3D scenes. Recent works have extended this method to dynamic 3D scenes by tracking dense scene elements~\cite{luiten2024dynamic} or modeling deformation fields~\cite{wu20244d, yang2024deformable}, enabling applications in dynamic environments~\cite{luiten2024dynamic,yang2023real,yang2024deformable}. Another line of research~\cite{chen2024text,tang2023dreamgaussian,yi2023gaussiandreamer} integrates Gaussian Splatting with diffusion-based models to create high-quality 3D content. In the domain of open-vocabulary segmentation on 3D Gaussians, recent methods~\cite{shi2024language,zhou2024feature,qin2024langsplat,ye2025gaussian} have explored ways to map 2D semantics onto 3D representations. LangSplat~\cite{qin2024langsplat} utilizes CLIP and SAM to project 2D semantic information onto 3D Gaussians, while Gaussian Grouping~\cite{ye2025gaussian} aligns SAM-generated masks across multiple views for consistent multi-view segmentation. However, these approaches are typically limited to specific trained scenes. In contrast, OVGaussian, trained on the SegGaussian dataset, enables open-vocabulary querying by directly predicting semantic property for each Gaussian in novel scenes, achieving cross-scene semantic generalization.

\section{Methodology}
\label{sec:methodology}

In this section, we present \textbf{OVGaussian}, a novel approach that enables 3D Gaussian representations \cite{kerbl20233d} with open-vocabulary segmentation capabilities across diverse scenes. Our method (Fig. \ref{fig:framework}) involves training a neural network on a large collection of Gaussian-based scenes to learn semantic representations for each Gaussian point. Once trained, the network can predict a semantic vector for each Gaussian point in a new, unseen Gaussian-based scene, enabling cross-scene generalization. This semantic vector can be rendered into open-vocabulary semantic maps from various viewpoints, allowing flexible, multi-view scene understanding. We begin by introducing \textbf{3D Gaussian splatting} as the core representation method for scenes, which serves as a basis for our segmentation approach. We then outline the primary property of OVGaussian: \textbf{Generalizable Semantic Rasterization (GSR)}, which allows 3D Gaussians to represent semantic information consistently across scenes, and \textbf{Cross-modal Consistency Learning (CCL)}, which ensures alignment between 3D and 2D semantic information for coherent, open-vocabulary segmentation.

\subsection{Preliminary of 3D Gaussian Splatting}

\textbf{3D Gaussian splatting} is a rendering technique that represents 3D scenes using a set of Gaussian functions distributed throughout the scene. Unlike traditional point clouds or mesh representations, 3D Gaussian splatting provides a continuous representation of spatial and semantic information, allowing for flexible and efficient rendering of complex 3D scenes. This representation supports multi-view rendering and allows for high-quality visualization of scenes from arbitrary viewpoints.

Each 3D Gaussian is parameterized by:

\begin{itemize}

\item {\textbf{Position \( p = (x, y, z) \):} The 3D coordinates representing the center of the Gaussian.}
\item {\textbf{Covariance matrix \( \Sigma \):} Determines the shape, spread, and orientation of the Gaussian in 3D space.}
\item {\textbf{Color \( c = (r, g, b) \):} The RGB color is associated with the Gaussian, contributing to the appearance of the scene.}
\item {\textbf{Opacity \( \alpha \):} Controls the transparency of the Gaussian, which influences how the Gaussian blends with others in the scene.}

\end{itemize}

The function for a 3D Gaussian can be expressed as:
\begin{equation}\label{equ:Rendering}
G(p; \Sigma) = \exp\left(-\frac{1}{2}(p - \mu)^{\top} \Sigma^{-1} (p - \mu)\right),
\end{equation}
where \( \mu \) is the Gaussian’s center, and \( \Sigma \) encodes its spread and orientation. During rendering, each Gaussian’s contribution to a pixel in the final image is weighted by its opacity and blended with neighboring Gaussians through alpha blending:
\begin{equation}\label{equ:Rendering_RGB}
\mathcal{I}(x, y) = \sum_{i=1}^N \alpha_i \, c_i \, G_i(x, y),
\end{equation}
where \( \mathcal{I}(x, y) \) is the pixel intensity at position \((x, y)\) on the rendered image, \( c_i \) and \( \alpha_i \) are the color and opacity of Gaussian \( G_i \). This process, known as \textbf{splatting}, allows efficient rendering of complex scenes with continuous, smooth representations from any viewpoint.

\subsection{Generalizable Semantic Rasterization}

To enable cross-scene semantic generalization, we introduce Generalizable Semantic Rasterization (GSR), which augments each 3D Gaussian with a semantic vector that carries consistent semantic information across scenes. GSR employs a multi-granularity fusion 3D neural network to predict this semantic vector for each Gaussian, facilitating multi-view consistent semantic rendering and establishing a shared semantic space across diverse scenes.

\vspace{-3ex}
\paragraph{Multi-granularity 3D Neural Network for 3D Semantic Vector Learning.} To efficiently predict the semantic vector for each Gaussian, we design a multi-granularity fusion 3D neural network \cite{choy20194d} that captures 3D spatial context across multiple granularities of the Gaussian point cloud. Our approach consists of two primary steps: (1) Voxelization and Sparse 3D Feature Extraction and (2) Voxel-to-Point Adapter.

Formally, let \( G = \{g_i\}_{i=1}^N \) represent the set of 3D Gaussians, where each Gaussian \( g_i \) is characterized by a position \( p_i \) and its semantic vector \( s_i \). The voxelization process produces a 3D grid of voxels \( V = \{v_j\} \), each voxel aggregating features from nearby Gaussians. The sparse 3D neural network \( F_{\text{s}} \) computes voxel-level features:
\begin{equation}\label{equ:sparse}
 \{f_{\text{voxel}}(v_j)\}_{j=1}^{|V|} =F_{\text{s}}(V) ,
\end{equation}
where \( f_{\text{voxel}}(v_j) \) denotes the feature of voxel \( v_j \).

\begin{comment}
    
Subsequently, the Voxel-to-Point Adapter interpolates these voxel features to derive a refined semantic vector for each Gaussian. The final semantic vector \( s_i \) for Gaussian \( g_i \) is computed by mapping the voxel features back to the point level features:
\begin{equation}\label{equ:Adapter}
s_i = \text{Adapter}(f_{\text{voxel}}(v_j), g_i),
\end{equation}
where \( \text{Adapter}(\cdot) \) is an attention-based neural network that converts voxel-level features of $g_{i}$ into Gaussian-level semantic vectors. This multi-granularity architecture enables GSR to capture both global and local 3D spatial information effectively, producing consistent and detailed semantic vectors for each Gaussian.
\end{comment}

\begin{figure}
  \centerline{\includegraphics[width=\linewidth]{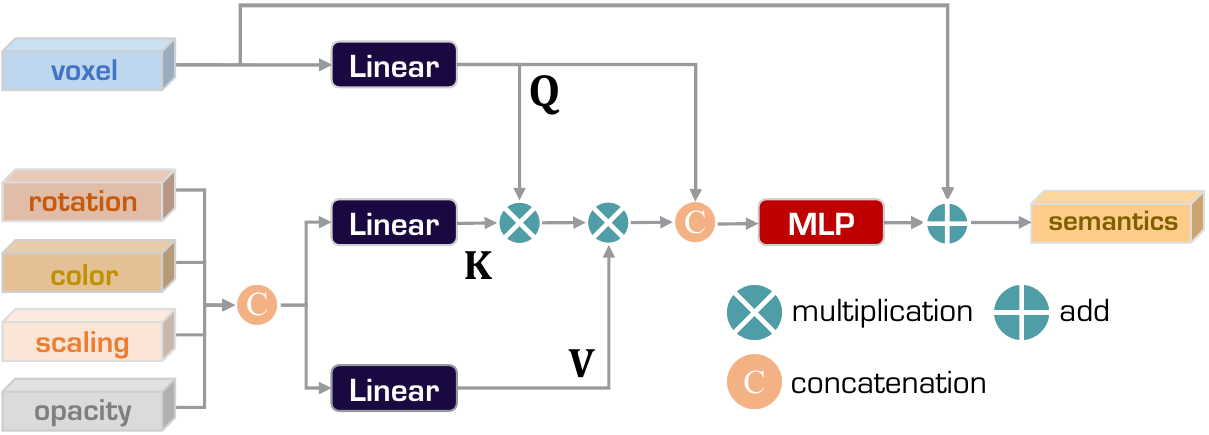}}
  \vspace{-0.2cm}
  \caption{Illustration of the adapter network. The adapter network takes voxel features and Gaussian attributes (rotation, color, scaling, and opacity) as inputs, processes them through a series of linear transformations and attention-based operations, and outputs a refined semantic vector for each Gaussian. This network enables effective multi-granularity fusion, capturing both local and global semantic information for each Gaussian point.}
  \label{fig:adapter}
  \vspace{-1ex}
\end{figure}

Subsequently, the Voxel-to-Point Adapter (Fig.~\ref{fig:adapter}) interpolates these voxel features to derive a refined semantic vector for each Gaussian. The final semantic vector \( s_i \) for Gaussian \( g_i \) is obtained by mapping the voxel features back to the point level using a mapping function $T_p(\cdot)$. The adapter mechanism is formulated as follows:
\begin{equation}
% \centering
\begin{aligned}
    m_i &= \text{MHSA}(\textbf{Q}(T_p(f_{\text{voxel}}(v_j))),\textbf{K}(T_p(g_i)),\textbf{V}(T_p(g_i)))~,
    \\
    s_{i} &= \text{MLP}(\text{Concat}(m_{i},T_p(f_{\text{voxel}}(v_j))))+T_p(f_{\text{voxel}}(v_j)))~,
\end{aligned}
\end{equation}
where $\textbf{Q}(\cdot)$, $\textbf{K}(\cdot)$, $\textbf{V}(\cdot)$ are linear projection layers that generate query, key, and value features, respectively. MHSA \cite{vaswani2017attention} denotes the multi-head self-attention module. The Voxel-to-Point Adapter employs attention mechanisms to integrate voxel-level features of $g_{i}$ into Gaussian-level semantic vectors. This multi-granularity design allows the GSR to effectively capture 3D spatial information at both global and local levels, producing consistent and detailed semantic vectors for each Gaussian.

\vspace{-3ex}
\paragraph{View-Independent Semantic Vector Representation.}
During training, we optimize each semantic vector in a manner similar to using Spherical Harmonic (SH) coefficients to represent color in 3D Gaussian-based appearance models. However, unlike view-dependent appearance modeling where SH coefficients change based on viewpoint, our semantic vector is designed to be \textbf{view-independent}. This consistency is essential for achieving a stable semantic representation of the scene across different viewpoints. Following Gaussian Grouping \cite{ye2025gaussian}, to enforce view-independence, we set the SH degree of the semantic vector to 0, thereby modeling only its direct-current (DC) component. This setup ensures that each Gaussian's semantic vector reflects a constant semantic identity, unaffected by viewing angle variations. 

\vspace{-3ex}
\paragraph{Multi-view Consistent 2D Semantic Rasterization.}
To render the learned semantic vectors as 2D semantic maps, we perform alpha blending, similar to rendering color property. These 2D maps provide a multi-view representation of the scene, with each Gaussian retaining both spatial and semantic information that remains robust across different scenes.

For a given viewpoint, let \( S = \{s_i\}_{i=1}^N \) denote the set of semantic vectors for all Gaussians. The rendered semantic map \( M(x, y) \) at pixel \((x, y)\) is obtained by blending the semantic contributions of each Gaussian in view:
\begin{equation}\label{equ:semantic}
\mathcal{M}(x, y) = \sum_{i=1}^N \alpha_i \, s_i \, G_i(x, y),
\end{equation}
where \( \alpha_i \) represents the opacity of Gaussian \( g_i \) and \( G_i(x, y) \) denotes its spatial distribution. This rendering process ensures that each Gaussian maintains consistent semantic information across views, enhancing the model's ability to generalize effectively across different scenes.

\subsection{Cross-modal Consistency Learning}

To incorporate open-vocabulary segmentation, we introduce Cross-modal Consistency Learning (CCL), which aligns the semantic information of 3D Gaussians with text embeddings and 2D image features. This alignment promotes a unified semantic understanding across modalities, enabling the model to interpret open-vocabulary terms consistently in 3D scenes.

\vspace{-3ex}
\paragraph{Semantic Alignment with Text Embeddings.}

To equip the 3D Gaussians with open-vocabulary segmentation capabilities, we align the semantic vectors of 3D Gaussians and their rendered 2D semantic maps with text embeddings, facilitating consistent and coherent open-vocabulary segmentation across scenes.

Each 3D Gaussian \( g_i \) in the scene is associated with a semantic label \( y_i \), which we map to an open-vocabulary embedding space using pre-trained embeddings, such as those from CLIP. Let \( E = \{e_m\}_{m=1}^M \) denote the set of text embeddings, where \( e_m \) corresponds to a semantic label in the open-vocabulary space.

To align the semantics, we use a cross-entropy loss that jointly optimizes the semantic alignment of both the 3D Gaussian semantic vectors \( s_i \) and the rendered 2D semantic map \( \mathcal{M}(x, y) \) of the scene. The loss function encourages each 3D Gaussian’s semantic vector and its corresponding 2D semantic map to be aligned with the correct text embedding.
\begin{align}\label{equ:Adapter}
\mathcal{L}_{\text{semantic}} = - \sum_{i=1}^N \underbrace{\log \frac{\exp(e_{\gamma_i}^\top \phi(s_i))}{\sum_{m=1}^M \exp(e_m^\top \phi(s_i))}}_{\text{3D-to-Text Alignment}} \\
- \sum_{(x,y) \in \mathcal{P}} \underbrace{\log \frac{\exp(e_{\gamma_{(x,y)}}^\top \phi(\mathcal{M}(x, y)))}{\sum_{m=1}^M \exp(e_m^\top \phi(\mathcal{M}(x, y)))}}_{\text{2D-to-Text Alignment}},
\end{align}
where \( \phi(\cdot) \) is a decoder network. \( N \) is the number of Gaussians in the scene, \( \mathcal{P} \) represents the set of pixels in the rendered 2D semantic map \( \mathcal{M} \). \( s_i \) is the semantic vector of Gaussian \( g_i \), and \( \mathcal{M}(x, y) \) denotes the semantic intensity at pixel \((x, y)\) in the rendered 2D map. \( e_{\gamma_i} \) and \( e_{(x,y))} \) are the text embedding corresponding to the semantic label of Gaussian \( g_i \) and pixel \((x, y)\), respectively.

By simultaneously aligning the 3D semantic vectors and 2D rendered maps with text embeddings, this cross-entropy loss enforces a unified semantic understanding that bridges 3D and 2D representations, enhancing the model’s ability to generalize across scenes and recognize novel categories in open-vocabulary segmentation tasks.

\begin{table*}[t]
  \centering
  \caption{Comparison of OVGaussian with baseline methods. Performance on 3D and 2D segmentation tasks measured by Cross-scene Accuracy (CSA), Open-vocabulary Accuracy (OVA), Novel View Accuracy (NVA), and Cross-Domain Accuracy (CDA), demonstrating OVGaussian’s superiority in accuracy and generalization.}
  \label{tab:comparison}
  \vspace{-0.2cm}
  \scriptsize
  \renewcommand{\arraystretch}{1}
  \centering
  \resizebox{\linewidth}{!}{
  \begin{tabular}{c |c | c c c | c c c c}
  \toprule
  \multirow{2}*{Methods} & \multirow{2}*{Publication}&
  \multicolumn{3}{c|}{3D (mIoU)} & \multicolumn{4}{c}{2D (mIoU)} \\
  % \cline{2-3}
  % \cline{4-5}
  % \cline{6-7}
  ~ && CSA & OVA & CDA & CSA & OVA & NVA & CDA
  \\\midrule	
  OpenScene~\cite{peng2023openscene}& CVPR 2023 & $30.22$ & $11.74$ & 10.22 & $36.18$ & $12.58$ & $52.20$ & 11.14
  \\ 	
  CLIP2Scene~\cite{chen2023clip2scene}& CVPR 2023 & $31.16$ & $11.98$ & 10.45 & $35.47$ & $12.77$ & $51.06$ & 11.32
  \\	
  CNS~\cite{chen2024towards}& NeurIPS 2023 & $36.21$ & $13.03$ & 12.64 & $40.78$ & $13.36$ & $59.28$ & 13.48
  \\	
  LangSplat~\cite{qin2024langsplat}& CVPR 2024& $21.23$ & $12.46$ & -& $25.67$ & $13.39$ & $41.26$ & -
  \\  
  Gaussian Grouping~\cite{ye2025gaussian}& ECCV 2024& $33.45$ & $12.01$ & -& $38.94$ & $13.03$ & $55.82$ & -

  \\\midrule 	
  Ours & -- &$\mathbf{43.84}$ & $\mathbf{15.24}$ & $\mathbf{18.93}$  & $\mathbf{45.76}$ & $\mathbf{16.27}$ &  $\mathbf{69.51}$ &$\mathbf{20.31}$ 
  \\\bottomrule
  \end{tabular}}
  \vspace{-1.5ex}
\end{table*}

\vspace{-3ex}
\paragraph{Dense Visual-Semantic Alignment.}
To enhance the open-vocabulary capabilities of 3D Gaussians, we incorporate Dense Visual-Semantic Alignment by leveraging pixel-level semantic information from a pre-trained 2D vision-language model. Specifically, we use MaskCLIP \cite{zhou2022maskclip} to extract dense pixel-level semantics from the original multi-view images, and we align these dense semantics with the corresponding pixels in the rendered 2D semantic maps, promoting a consistent semantic representation across 2D and 3D modalities.

To enforce this dense visual-semantic consistency, we employ a cosine similarity loss that aligns each pixel in the 2D semantic map with the corresponding pixel in the MaskCLIP-extracted dense semantics. This approach allows the model to learn pixel-level semantics directly from the pre-trained 2D model, enriching the open-vocabulary segmentation capability of the 3D Gaussians.

The cosine similarity loss is defined as:
\begin{equation}\label{equ:cosine}
\mathcal{L}_{\text{cosine}} = - \frac{1}{|\mathcal{P}|} \sum_{(x, y) \in \mathcal{P}} \frac{\mathcal{S}(x, y) \cdot \psi(\mathcal{M}(x, y))}{\|\mathcal{S}(x, y)\| \|\psi(\mathcal{M}(x, y))\|},
\end{equation}
where \( \psi(\cdot) \) is a decoder network. \( \mathcal{P} \) represents the set of pixels in the image. \( \mathcal{S}(x, y) \) is the dense semantic embedding from MaskCLIP at pixel \((x, y)\). \( \mathcal{M}(x, y) \) is the rendered semantic embedding from the 2D semantic map at the same pixel.

This cosine similarity loss encourages each pixel in the rendered 2D semantic map to align closely with the corresponding dense semantic representation obtained from MaskCLIP. By enforcing dense alignment at the pixel level, the model can learn detailed open-vocabulary semantics directly from the 2D large model, further enhancing its ability to generalize across diverse scenes and recognize novel categories in open-vocabulary segmentation tasks.

\subsection{Training and Inference with SegGaussian}

We train OVGaussian using the SegGaussian dataset, which includes 288 scenes represented as 3D Gaussians with semantic and instance labels for both Gaussian points and multi-view images. This multimodal dataset enables the model to learn robust cross-modal representations, enhancing generalization to new scenes and domains.

During training, we jointly optimize the GSR and CCL components. The total loss function is defined as:
\begin{equation}\label{equ:loss}
\mathcal{L} = \mathcal{L}_{\text{semantic}} + \mathcal{L}_{\text{cosine}},
\end{equation}
where \( \mathcal{L}_{\text{semantic}} \) aligns 3D semantic representations with text embeddings, and \( \mathcal{L}_{\text{cosine}} \) enforces consistency across 2D and 3D modalities. This joint optimization encourages the model to learn semantic information that is robust across modalities and generalizable across diverse 3D scenes.

At inference time, the model is guided by a set of text embeddings, with unseen Gaussian-based scenes serving as queries, allowing OVGaussian to perform open-vocabulary segmentation without additional fine-tuning.

\section{Experiments}
\label{sec:experiments}

\begin{figure*}
  %%%%\vspace*{-5ex}
  \centerline{\includegraphics[width=1\textwidth]{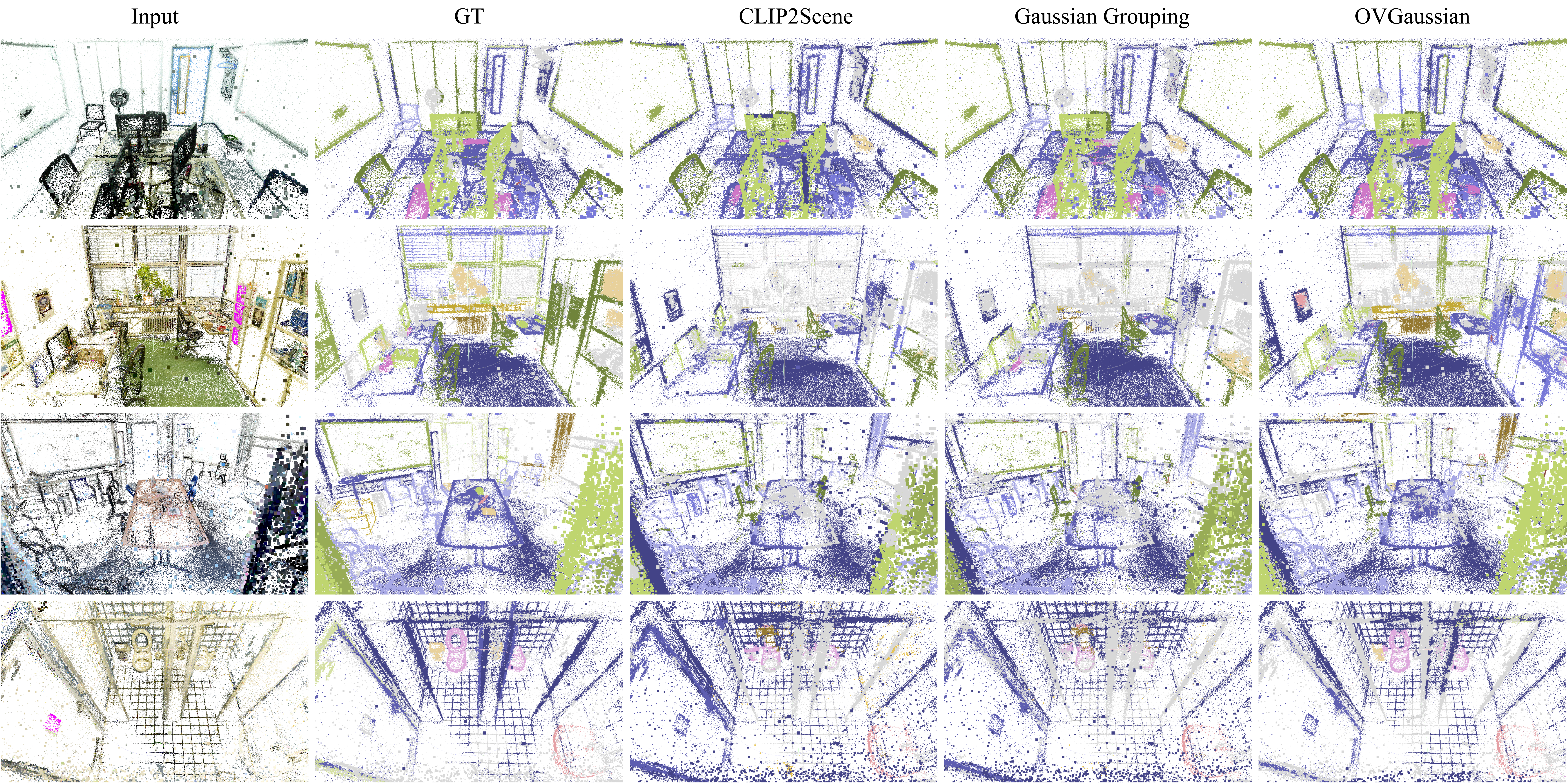}}
  %\vspace*{-0.5ex}
  \vspace{-1.5ex}
  \caption{Quantitative comparisons of 3D Cross-Scene Accuracy (CSA) across different methods: CLIP2Scene, Gaussian Grouping, and OVGaussian. The figure highlights our enhanced segmentation accuracy and consistency, especially in handling complex scene details.}
  \label{fig:visual_3D}
  \vspace{-1ex}
\end{figure*}

In this section, we evaluate \textbf{OVGaussian} on its ability to perform open-vocabulary segmentation across diverse 3D scenes. We first provide an overview of the \textbf{SegGaussian} dataset and the experimental setup, including baseline methods and evaluation metrics. We then present comprehensive quantitative and qualitative results, demonstrating OVGaussian’s effectiveness in cross-scene, cross-domain, and novel-view generalization. Additionally, we perform ablation studies to analyze the contributions of OVGaussian’s key components, \textbf{Generalizable Semantic Rasterization (GSR)} and \textbf{Cross-modal Consistency Learning (CCL)}, and provide a detailed efficiency analysis.

\subsection{SegGaussian Dataset}
The \textbf{SegGaussian} dataset is constructed from two well-established datasets: \textbf{ScanNet++} \cite{yeshwanth2023scannet++} and \textbf{Replica} \cite{straub2019replica}. Specifically, SegGaussian comprises 280 scenes from ScanNet++ and 8 scenes from Replica, resulting in a total of 288 scenes. We split the dataset into the training, validation and cross-domain validation set, with 230, 50 and 8 scenes, respectively. Both ScanNet++ and Replica provide detailed 3D point clouds, multi-view RGB images, and corresponding camera poses. Besides, semantic and instance annotations for point clouds and images are also available, making them ideal sources for building a comprehensive dataset suited to open-vocabulary 3D segmentation.

To represent each scene as a Gaussian-based model, we use 3D Gaussian splatting \cite{kerbl20233d} to convert posed images into 3D Gaussian representation. Each scene’s 3D Gaussian model captures both spatial structure and semantic context, supporting high-quality rendering and segmentation from multiple viewpoints.

For each 3D Gaussian in a scene, we assign semantic and instance labels by aligning the 3D Gaussian points with the annotations available in the 3D point clouds from ScanNet++ and Replica. This labeling process ensures that each Gaussian is enriched with detailed semantic and instance information, which can be used for both 3D segmentation and rendering of semantic maps in various views. This semantic annotation setup provides a robust foundation for training and evaluating models on open-vocabulary 3D segmentation tasks, enabling detailed analysis of cross-scene generalization, open-vocabulary recognition, and multi-view consistency. Details are in the supplementary materials.

\subsection{Experimental Setup}

\paragraph{Baseline Settings.}
We compare OVGaussian against several state-of-the-art methods adapted for open-vocabulary segmentation in 3D scenes:
1). \textbf{OpenScene \cite{peng2023openscene}}: This method adapts the 2D vision model by lifting 2D image features to 3D points, attempting to preserve the semantic richness of 2D models in 3D space.
2). \textbf{CLIP2Scene \cite{chen2023clip2scene}}: This baseline combines MaskCLIP for dense pixel semantics with a 3D PointNet that maps the 2D semantics to 3D points.
3). \textbf{LangSplat \cite{qin2024langsplat}}: This baseline combines CLIP and SAM for mapping the 2D semantics to 3D Gaussians.
4). \textbf{Gaussian Grouping \cite{ye2025gaussian}}: A recent method that aligns SAM's predicted masks across multiple views, promoting consistent multi-view segmentation. \textbf{OpenScene}, \textbf{CLIP2Scene} represent various strategies for transferring 2D open-vocabulary knowledge to 3D spaces, but they lack the Gaussian-based representation and multi-modal consistency mechanisms of OVGaussian. For a fair comparison, we use the 3D points GT during the training stage.
\textbf{LangSplat} and \textbf{Gaussian Grouping} are scene-specific methods that could not transfer the semantic query across scenes. For these two methods, we train all the test scenes for comparison.

\vspace{-3ex}
\paragraph{Evaluation Metrics.}
We evaluate models using four primary metrics, all measured by Mean Intersection-over-Union (mIoU):
\textbf{Cross-scene Accuracy (CSA)}: Measures the accuracy of predicted segmentations across all semantic classes in the test scenes, assessing the model’s ability to segment objects at the scene level.
\textbf{Open-vocabulary Accuracy (OVA)}: Measures the model’s segmentation accuracy on categories that are not seen during training, testing open-vocabulary generalization.
\textbf{Novel View Accuracy (NVA)}: Quantifies the segmentations performance of the novel views in the training scenes. \textbf{Cross-Domain Accuracy (CDA)}: Indicates the segmentations performance across different data domains, here we use the 8 scenes from the Replica dataset for evaluation.

\vspace{-3ex}
\paragraph{Implementation Details.}
We employ MinkowskiNet34C~\cite{choy20194d} as the 3D backbone network. The model is trained using an SGD optimizer with a learning rate of 0.02 and a batch size of 3. For efficient training, each 3D Gaussian scene is paired with a single-view image. Training the model for 300 epochs on a single H100 GPU takes approximately 20 hours.

\begin{figure*}
  %%%%\vspace*{-5ex}
  \centerline{\includegraphics[width=1\textwidth]{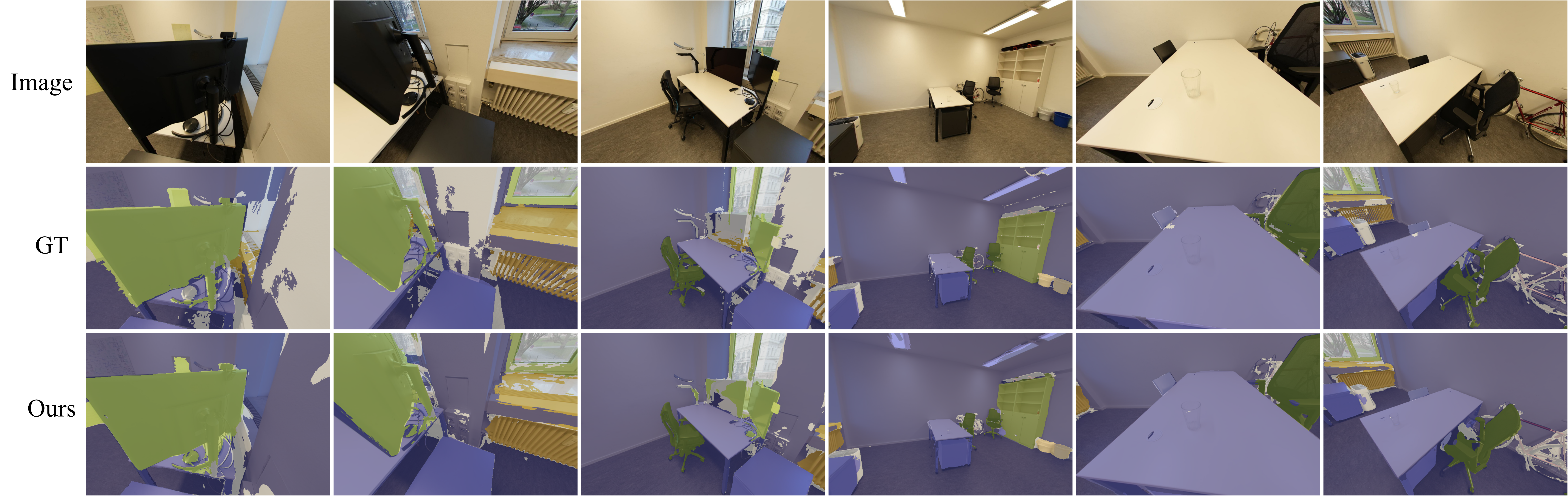}}
  %\vspace*{-0.5ex}
  \vspace{-1.5ex}
  \caption{Visualization of cross-view consistency in novel viewpoints. It illustrates OVGaussian’s ability to maintain coherent segmentation across diverse viewpoints, showcasing cross-view consistency in 3D scene understanding.}
  \label{fig:visual_2D}
  \vspace{-1ex}
\end{figure*}

\subsection{Comparison Results and Discussion}

\begin{table*}[t]
  \centering
  \caption{Evaluations of the impact of Visual-Semantic alignment (V-S align) and 3D-to-Text alignment (3D-T align) from Cross-Modal Consistency Learning (CCL) and the adapter from Generalizable Semantic Rasterization (GSR). Results show performance on 3D CSA, 2D CSA, and NVA across $20$, $50$, and $100$ categories, highlighting each component’s effect to improve segmentation accuracy and consistency.}
  \label{tab:ablation}
  \vspace{-0.3cm}
  \footnotesize
  \renewcommand{\arraystretch}{1}
  \centering
  \resizebox{\linewidth}{!}{
  \begin{tabular}{c c c |c c c | c c c | c c c}
  \toprule
  \multicolumn{3}{c|}{Settings} & 
  \multicolumn{3}{c|}{3D CSA} & \multicolumn{3}{c|}{2D CSA} & \multicolumn{3}{c}{NVA} \\
  % \cline{2-3}
  % \cline{4-5}
  % \cline{6-7}
  V-S align & 3D-T align & adapter & $20$& $50$ & $100$ & $20$ & $50$ & $100$ & $20$ & $50$ & $100$
  \\\midrule
  & & & $32.18$  & $20.34$ & $15.26$ & $37.15$  & $24.58$ & $17.24$ & $56.35$  & $36.31$ & $18.42$
  \\	
  \checkmark & & & $34.57$  & $23.57$ & $15.66$ & $40.17$  & $26.73$ & $18.34$ & $57.71$  & $37.93$ & $19.23$
  \\	
  \checkmark&\checkmark&& $40.55$  & $30.39$ & $19.11$ & $43.70$  & $31.73$ & $20.85$ & $67.55$  & $53.71$ & $27.98$
  \\	
  \checkmark&\checkmark&\checkmark &$\mathbf{43.84}$ & $\mathbf{33.84}$ &  $\mathbf{20.73}$ &$\mathbf{45.76}$ & $\mathbf{33.76}$ &  $\mathbf{21.84}$ &$\mathbf{69.51}$ & $\mathbf{55.35}$ &  $\mathbf{28.85}$ 
  \\\bottomrule
  \end{tabular}}
  \vspace{-3ex}
\end{table*}

\begin{comment}
    
\vspace{-3ex}
\paragraph{Novel View Accuracy (NVA).}
OVGaussian achieves a high NVA of 69.51\% in 2D, significantly outperforming the baseline CNS, which achieves 59.28\%. This result highlights OVGaussian’s superior ability to maintain consistent segmentations across different viewpoints, a critical factor for robust 3D applications. The view-invariant semantic vectors provided by the GSR module enable OVGaussian to maintain stable semantic representations, regardless of the rendering angle, allowing for coherent segmentation across various viewpoints (Fig. \ref{fig:visual_2D}). This ability is particularly valuable in 3D applications where scenes are viewed from multiple perspectives.

\vspace{-3ex}
\paragraph{Cross-Domain Accuracy (CDA).}
On CDA, which tests the model’s adaptability to new domains, OVGaussian achieves 18.93\% in 3D and 20.31\% in 2D, outperforming CNS, which scores 12.64\% and 13.48\% in 3D and 2D, respectively. This substantial improvement demonstrates OVGaussian’s robustness in adapting to new domains, which is crucial for applications where a model may need to operate across different environments. The consistent Gaussian representation, combined with the semantic alignment achieved through CCL, enhances the model’s generalization ability across domains, enabling it to handle domain shifts more effectively than baseline methods.
\end{comment}

Table \ref{tab:comparison} compares the performance of OVGaussian with several state-of-the-art methods on 20 categories for open-vocabulary 3D Gaussian segmentation across four metrics: CSA, OVA, NVA, and CDA. OVGaussian consistently outperforms baselines in both 3D and 2D tasks, demonstrating strong generalization across scenes, domains, and viewpoints.
\vspace{-3ex}
\paragraph{Cross-scene Accuracy (CSA).}
OVGaussian achieves a CSA of 43.84\% in 3D and 45.76\% in 2D, surpassing CNS (36.21\% in 3D, 40.78\% in 2D). This reflects OVGaussian’s ability to segment objects consistently across diverse scenes, enabled by the \textbf{Generalizable Semantic Rasterization (GSR)} module, which ensures semantic consistency across scenes and viewpoints (Fig. \ref{fig:visual_3D}).
\vspace{-3ex}
\paragraph{Open-vocabulary Accuracy (OVA).}
With an OVA of 15.24\% in 3D and 16.27\% in 2D, OVGaussian outperforms CNS by over 2\%, demonstrating its strength in recognizing unseen categories. The \textbf{Cross-modal Consistency Learning (CCL)} module aligns 3D Gaussian semantics with open-vocabulary embeddings, enhancing recognition of novel categories.
\vspace{-3ex}
\paragraph{Novel View Accuracy (NVA).}
OVGaussian attains an NVA of 69.51\% in 2D, outperforming CNS (59.28\%). The view-invariant semantic vectors from GSR allow OVGaussian to maintain stable representations across viewpoints, crucial for coherent 3D segmentation (Fig. \ref{fig:visual_2D}).
\vspace{-3ex}
\paragraph{Cross-Domain Accuracy (CDA).}
On CDA, OVGaussian achieves 18.93\% in 3D and 20.31\% in 2D, outperforming CNS (12.64\% in 3D, 13.48\% in 2D). This highlights OVGaussian’s adaptability to new domains, supported by consistent Gaussian representation and CCL’s semantic alignment, enabling robust cross-domain generalization.

\subsection{Ablation Studies}
To analyze the contributions of the core components in OVGaussian, we conduct an ablation study on \textbf{Cross-modal Consistency Learning (CCL)} and \textbf{Generalizable Semantic Rasterization (GSR)}. In this study, we evaluate the effect of \textbf{Visual-semantic alignment (V-S align)} and \textbf{3D-to-text alignment (3D-T align)} (components of CCL) and the \textbf{adapter} module (a component of GSR). Table \ref{tab:ablation} shows the results, measured in terms of 3D CSA, 2D CSA, and NVA, across different numbers of categories: 20, 50, and 100.

\vspace{-3ex}
\paragraph{Effect of Cross-modal Consistency Learning (CCL).}
The V-S and 3D-T alignment components of CCL align 3D Gaussian semantic vectors with 2D image semantics and open-vocabulary embeddings, enhancing OVGaussian's open-vocabulary segmentation. \textbf{V-S alignment only}: Adding V-S alignment improves 3D CSA, 2D CSA, and NVA across all category counts compared to the baseline without alignment. For 100 categories, 3D CSA increases from 15.26\% to 15.66\%, and 2D CSA rises from 17.24\% to 18.34\%, indicating strengthened semantic consistency. \textbf{V-S + 3D-T alignment}: Adding 3D-T alignment further boosts NVA from 19.23\% to 27.98\%, enabling OVGaussian to better leverage open-vocabulary knowledge and generalize across novel classes and perspectives.

\vspace{-3ex}
\paragraph{Effect of Generalizable Semantic Rasterization (GSR).}
The \textbf{adapter} module in GSR refines semantic representations by transforming voxel features into fine-grained Gaussian features. This multi-granularity fusion improves cross-scene and open-vocabulary segmentation. \textbf{Adapter enabled}: Including the adapter improves metrics across the board. For 100 categories, 3D CSA rises from 19.11\% to 20.73\%, 2D CSA from 20.85\% to 21.84\%, and NVA reaches 28.85\%. The adapter enhances view consistency, capturing both global and detailed features for better semantic generalization across scenes.

\section{Conclusions}
\label{sec:conclusions}
We introduced SegGaussian, a dataset for open-vocabulary 3D segmentation with Gaussian-based representations. Building on this dataset, we developed OVGaussian, an algorithm enabling 3D Gaussians to perform open-vocabulary segmentation with strong generalization across scenes, domains, and viewpoints. OVGaussian integrates Generalizable Semantic Rasterization (GSR) for consistent semantic representations and Cross-modal Consistency Learning (CCL) to align 3D Gaussian semantics with 2D visual and text embeddings. Extensive experiments show OVGaussian’s state-of-the-art performance, underscoring its potential for versatile open-vocabulary 3D scene understanding.

\newcommand{\ourmodel}{\textit{OVGaussian}}

\def\paperID{522} % *** Enter the Paper ID here
\def\confName{CVPR}
\def\confYear{2025}

\renewcommand\thefigure{\Alph{figure}}
\renewcommand\thetable{\Alph{table}}
\renewcommand\thesection{\Alph{section}}

%\title{OVGaussian: Generalizable 3D Gaussian Segmentation with Open Vocabularies}

% Title
\maketitlesupplementary

% Table of Contents
\section*{Table of Contents}
\startcontents[appendices]
\printcontents[appendices]{l}{1}{\setcounter{tocdepth}{3}}

\section{The SegGaussian Dataset}
\label{sec:SegGaussian}
In this section, we elaborate on additional details of the data structure, construction procedures, statistics, and more examples of the proposed \textbf{SegGaussian} dataset.

\subsection{Dataset Overview}
\begin{figure}
  \centerline{\includegraphics[width=0.5\textwidth]{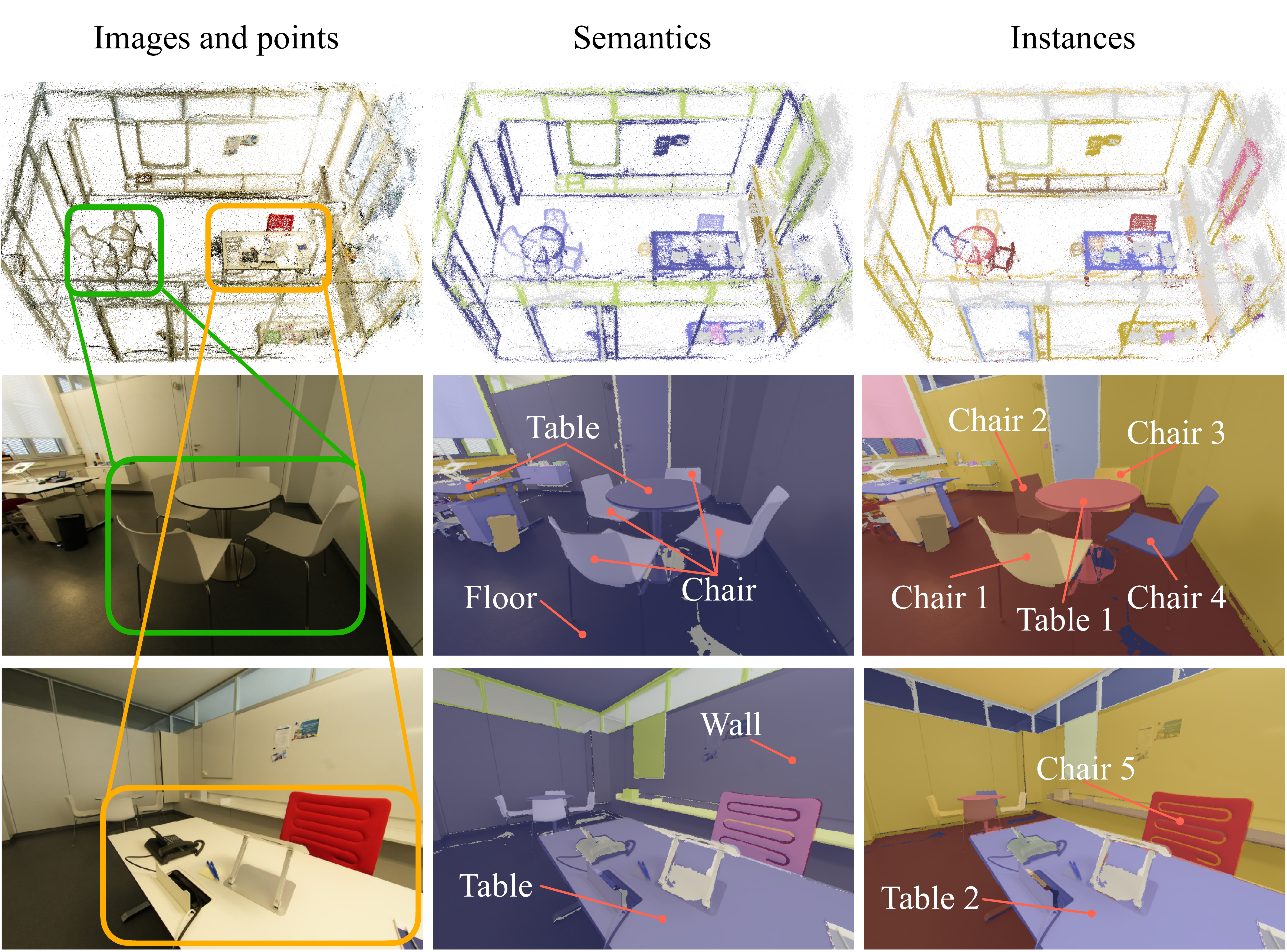}}
  \vspace{-1.2ex}
  \caption{Visualization of samples from the \textbf{SegGaussian} dataset, showcasing multi-view RGB images, semantic annotations (\eg, wall, floor, chair, table), and instance-level segmentation for 3D Gaussians. This comprehensive representation highlights the detailed semantic and instance information provided for each scene, supporting robust evaluation of open-vocabulary 3D segmentation.}
  \label{fig:segPoint_sample}
\end{figure}

The \textbf{SegGaussian} dataset is constructed from two well-established datasets: ScanNet++ \cite{yeshwanth2023scannet++} and Replica \cite{straub2019replica}. Specifically, SegGaussian comprises $280$ scenes from ScanNet++ and $8$ scenes from Replica, resulting in a total of $288$ scenes. We split the dataset into the training, validation, and cross-domain validation sets, with $230$, $50$, and $8$ scenes, respectively. Both ScanNet++ and Replica provide detailed 3D point clouds, multi-view RGB images, and corresponding camera poses. Besides, semantic and instance annotations for point clouds and images are also available, making them ideal sources for building a comprehensive dataset suited to open-vocabulary 3D segmentation (\cref{fig:segPoint_sample}).

\subsection{Dataset Construction}
To represent each scene as a Gaussian-based model, we use 3D Gaussian splatting \cite{kerbl20233d} to convert posed images into 3D Gaussian representation. On average, this conversion process takes approximately 50 minutes per scene on an NVIDIA H100 GPU. Each scene’s 3D Gaussian model captures both spatial structure and semantic context, supporting high-quality rendering and segmentation from multiple viewpoints.

For each 3D Gaussian in a scene, we assign semantic and instance labels by aligning the 3D Gaussian points with the annotations available in the 3D point clouds from ScanNet++ and Replica. This labeling process ensures that each Gaussian is enriched with detailed semantic and instance information, which can be used for both 3D segmentation and rendering of semantic maps in various views. For the evaluation of scenes from the Replica dataset, we use only the common categories shared between Replica and ScanNet++, ensuring consistency and comparability across datasets. This detailed annotation setup provides a robust foundation for training and evaluating models on open-vocabulary 3D segmentation tasks, enabling a comprehensive analysis of cross-scene generalization, open-vocabulary recognition, and multi-view consistency.

\begin{figure*}
  \centerline{\includegraphics[width=1\textwidth]{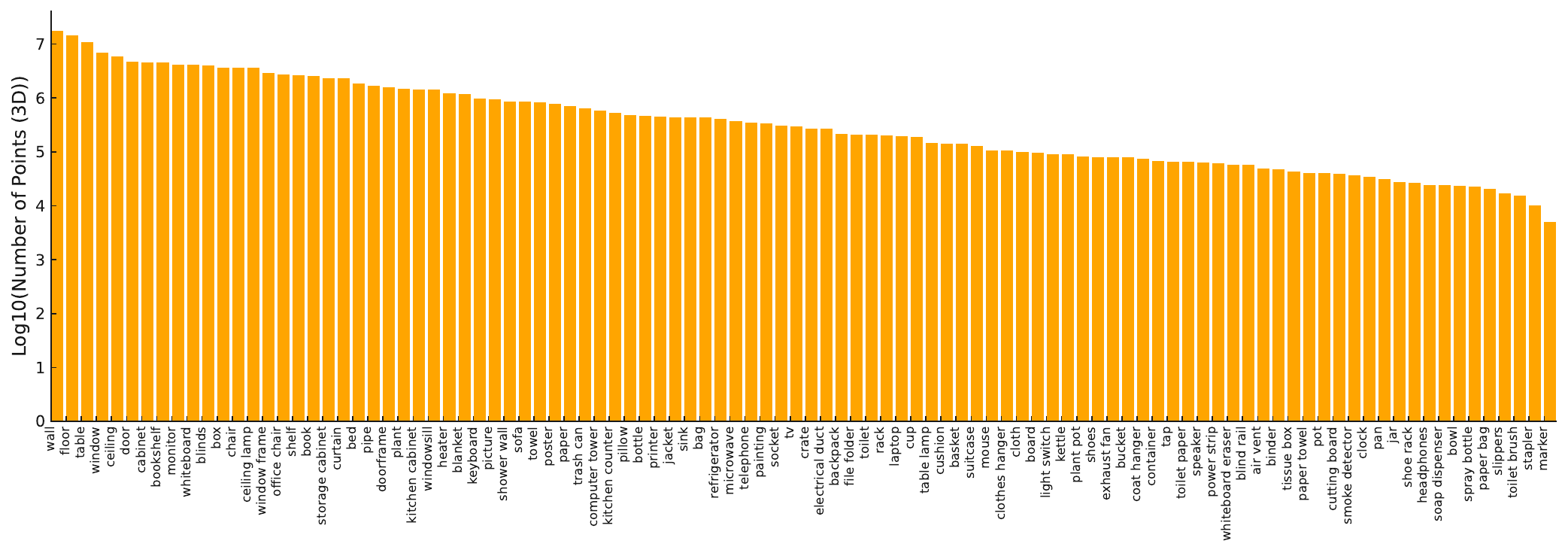}}
  \vspace{-2ex}
  \caption{Distribution of the top 100 semantic classes based on the logarithmic number of 3D points (log10 scale) in the dataset.}
  \vspace{-2ex}
  \label{fig:Top_100_Classes}
\end{figure*}

\begin{figure}
    \centerline{\includegraphics[width=0.5\textwidth]{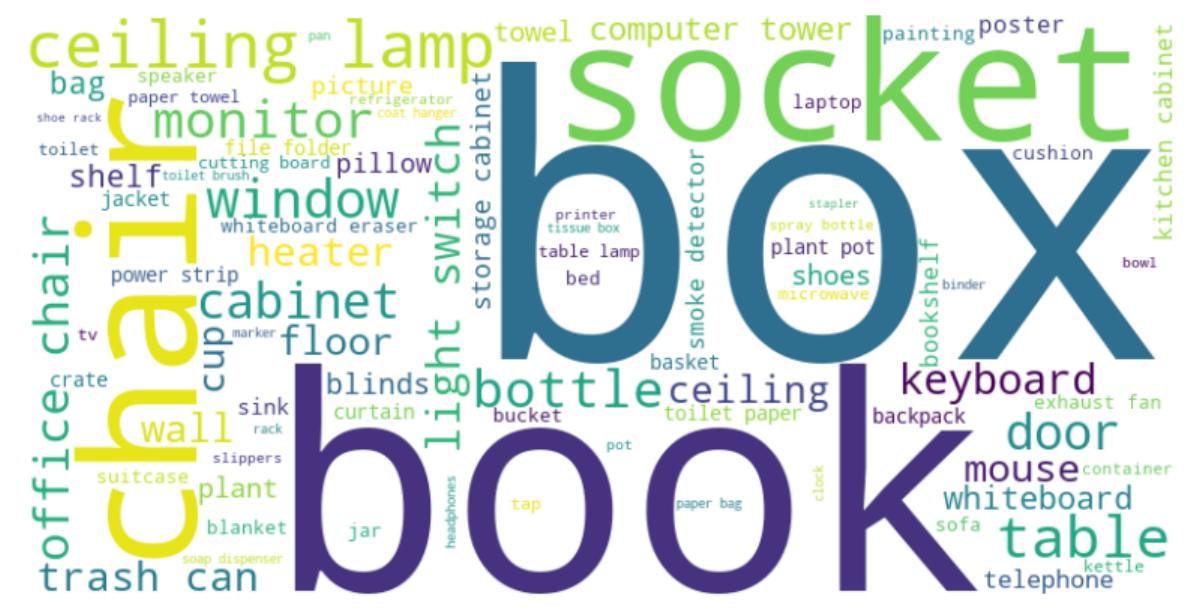}}
  \vspace{-1.2ex}
  \caption{The word cloud of top 100 classes based on the number of instances in the dataset.}
  \vspace{-2ex}
  \label{fig:Top_100_word_cloud}
  
\end{figure}

\subsection{Dataset Statistics}
Following the evaluation protocol of ScanNet++, the semantic categories consist of $100$ classes, including $97$ thing classes (\eg, objects such as tables, chairs, and books) and $3$ stuff classes (floor, ceiling, and walls). 

We illustrate the frequency distribution of these 100 categories in \cref{fig:Top_100_Classes}, providing a detailed view of the semantic class in the dataset. Additionally, in \cref{fig:Top_100_word_cloud}, we present a word cloud that visualizes the instance counts for each category, emphasizing the prevalence of different semantic classes based on their instance frequency. These visualizations provide valuable insights into the dataset’s composition, highlighting the diversity and density of semantic categories.
% \cref{fig:top_100_cls}

% \begin{figure*}
%   %%%%\vspace*{-5ex}
%   \centerline{\includegraphics[width=1\textwidth]{v0/figures/suplementary_100_classes.pdf}}
%   %\vspace*{-0.5ex}
%   \vspace{-1.5ex}
%   \caption{100 class}
%   \label{fig:top_100_cls}
% \end{figure*}

\begin{figure*}
  %%%%\vspace*{-5ex}
  \centerline{\includegraphics[width=1\textwidth]{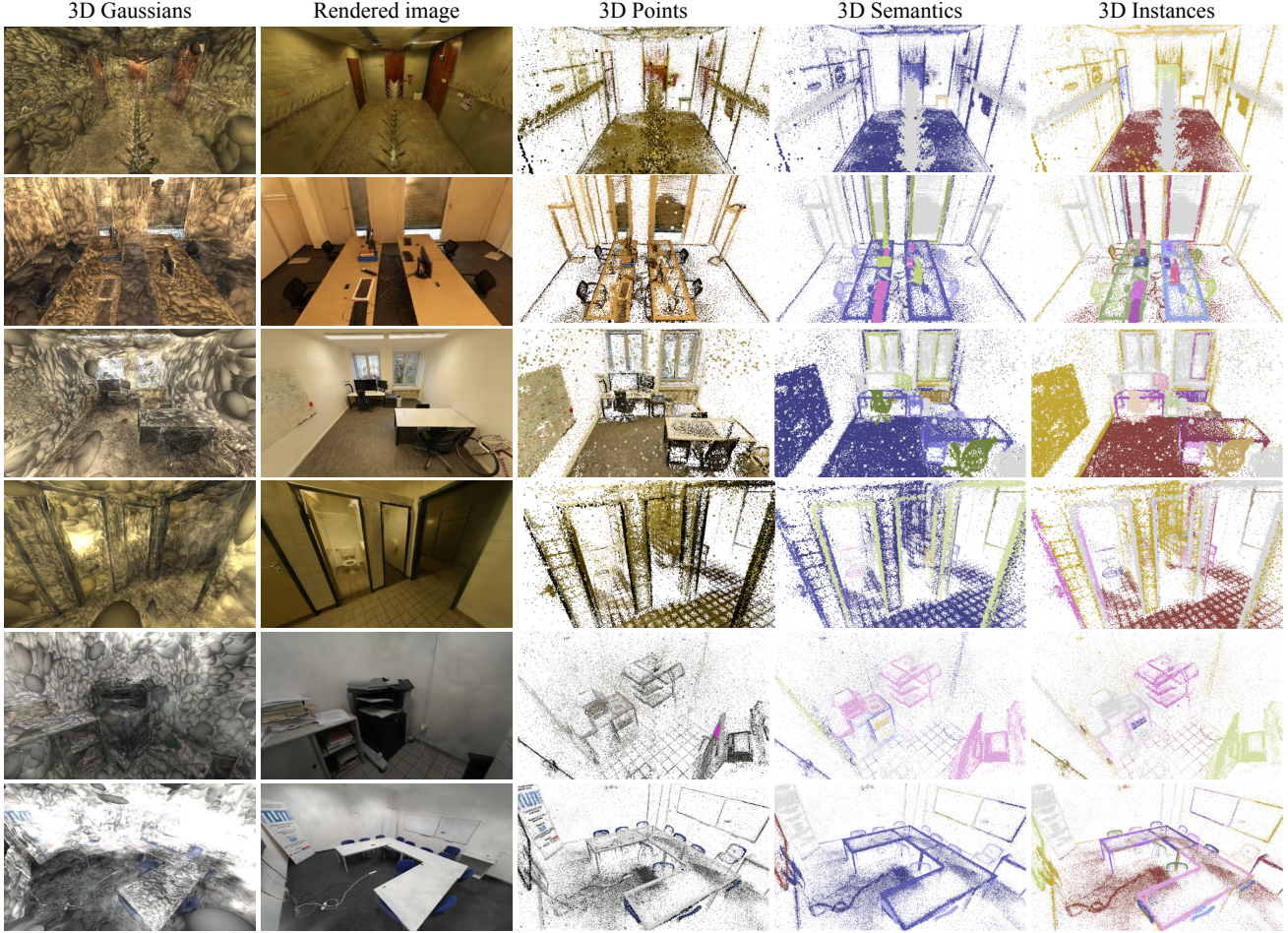}}
  %\vspace*{-0.5ex}
  \vspace{-1.2ex}
  \caption{Examples from the \textbf{SegGaussian} dataset. The figures showcase the 3D Gaussians, rendered images, 3D points, 3D semantic and instance annotations.}
  \label{fig:visual_segPoint_3D}
\end{figure*}

\subsection{Dataset Examples}
\begin{figure*}
  \centerline{\includegraphics[width=1\textwidth]{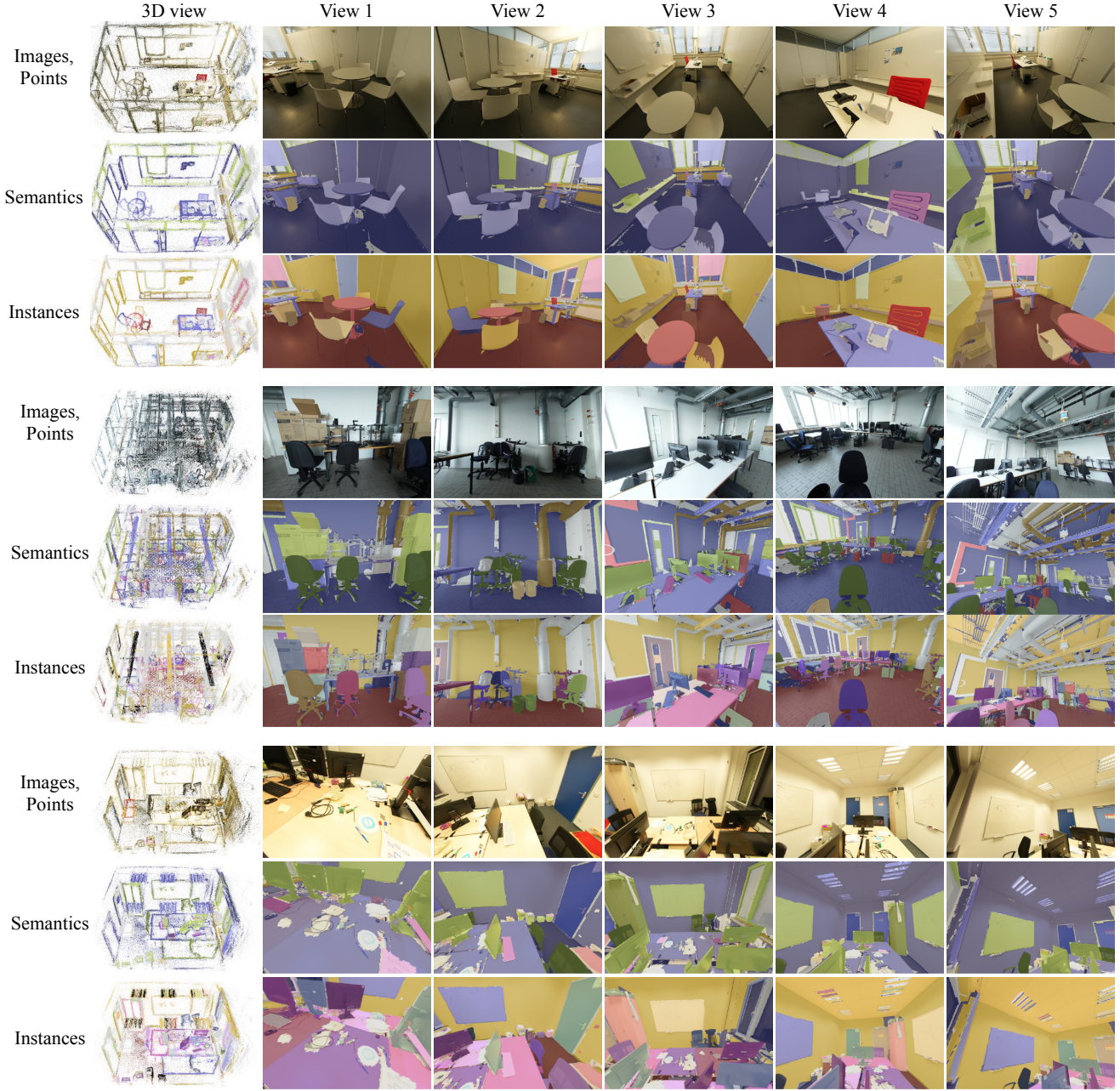}}
  %\vspace*{-0.5ex}
  \vspace{-1.2ex}
  \caption{Examples from the \textbf{SegGaussian} dataset. The figure shows multi-view consistent semantic and instance annotations.}
  \label{fig:visual_segPoint_2D}
\end{figure*}

The \textbf{SegGaussian dataset} provides comprehensive annotations for 3D scenes, incorporating semantic and instance labels for Gaussian-based representations (\cref{fig:segPoint_sample}, \cref{fig:visual_segPoint_3D}
 and \cref{fig:visual_segPoint_2D}). Each sample in the dataset consists of: 
 \begin{itemize}
     \item \textbf{RGB Images}: Multi-view images capturing the scene from different perspectives, serving as input for semantic and instance segmentation tasks. 

     \item \textbf{3D Gaussian Representation}: A Gaussian-based point cloud that represents the spatial and semantic structure of the scene, offering a continuous and efficient 3D representation. 

     \item \textbf{Semantic Annotations}: Each Gaussian point is assigned a semantic category (\eg, wall, floor, chair, table), enabling a detailed understanding of the scene's components. 

     \item  \textbf{Instance Annotations}: For object-centric tasks, Gaussian points are further labeled with instance identifiers (\eg, Chair 1, Chair 2, Table 1), distinguishing individual objects of the same category within the scene.
 \end{itemize}
This dataset supports the development and evaluation of models for open-vocabulary 3D segmentation, emphasizing cross-scene generalization, open-vocabulary recognition, and multi-view consistency.

\section{Additional Implementation Details}
\label{sec:details}
In this section, we provide additional details to facilitate the implementation and reproducibility of the proposed \textbf{OVGaussian} framework.

\subsection{Training Configurations}
We employ MinkowskiNet34C~\cite{choy20194d} as the 3D backbone network. The model is trained using the Stochastic Gradient Descent (SGD) optimizer with an initial learning rate of $0.02$ and a batch size of $3$. For efficient training, each 3D Gaussian scene is paired with a corresponding single-view image. To improve the robustness of the 3D neural network, we apply data augmentation techniques during training, including random scaling, random rotation, and flipping operations on the point cloud data. These augmentations help the model generalize better to variations in scene geometry and layout. We train the model for a total of $300$ epochs on a single NVIDIA H100 GPU, completing the process in approximately $20$ hours. To evaluate Open-vocabulary Accuracy (OVA), we set the unseen classes to be curtain, bookshelf, sofa, and bed.

\subsection{Gaussian Representations}
For each Gaussian, the semantic vector has a dimension of $16$, rotation is represented with $4$ dimensions, color with $3$ dimensions, scaling with $3$ dimensions, and opacity with $1$ dimension. The adapter module utilizes a multi-layer perceptron (MLP) with hidden layers of dimensions $27$, $96$, $96$, and $16$, each followed by a ReLU activation function. The decoder network \( \phi(\cdot) \) adopts a multi-layer perceptron (MLP) with hidden layers of dimensions $16$, $128$, and $512$, each followed by a ReLU activation function. The decoder network \( \psi(\cdot) \) adopts a multi-layer perceptron (MLP) with hidden layers of dimensions $16$, $128$, and $512$, each followed by a ReLU activation function. The image resolution used during training is \(584 \times 876\). This configuration enables effective learning while maintaining computational efficiency.

\section{Additional Experimental Results}
\label{sec:qualitative}
\begin{figure*}
  %%%%\vspace*{-5ex}
  \centerline{\includegraphics[width=1\textwidth]{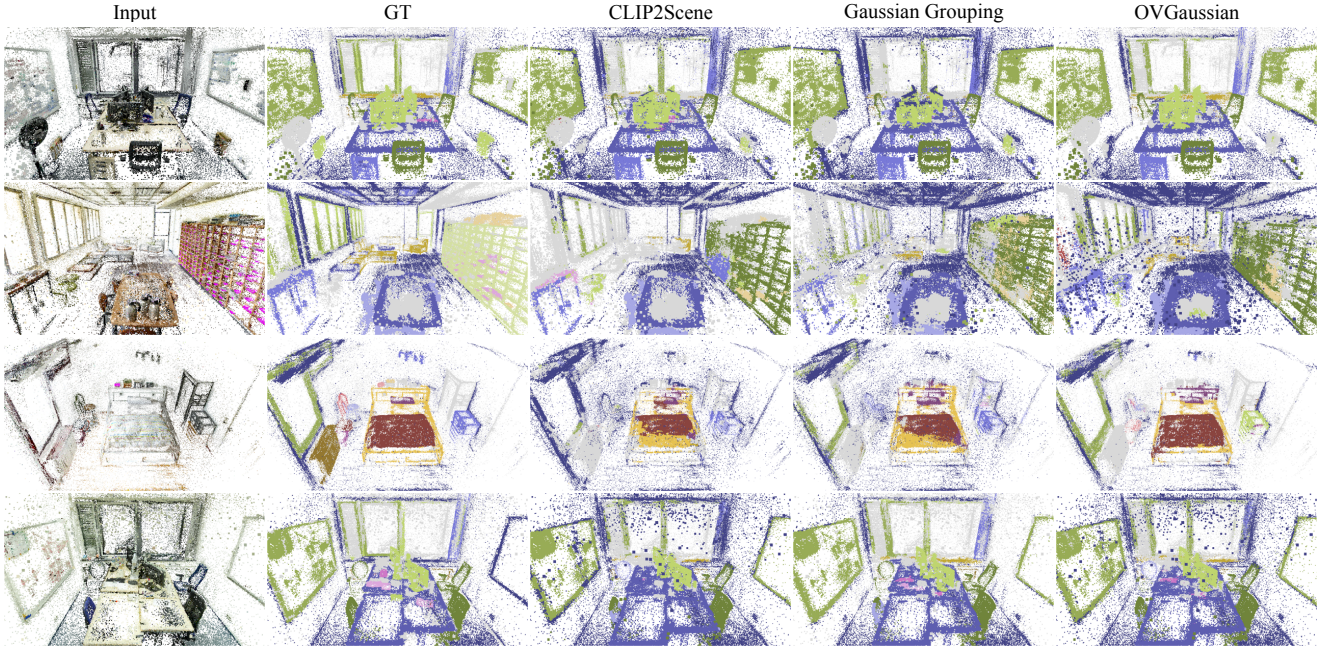}}
  %\vspace*{-0.5ex}
  \vspace{-1.2ex}
  \caption{Qualitative comparisons of 3D Cross-Scene Accuracy (CSA) across different methods: CLIP2Scene, Gaussian Grouping, and OVGaussian. The figure highlights our enhanced segmentation accuracy and consistency, especially in handling complex scene details.}
  \label{fig:visual_3D_sup}
\end{figure*}

We conducted a qualitative analysis to compare the segmentation performance of OVGaussian against state-of-the-art methods, including CLIP2Scene \cite{chen2023clip2scene} and Gaussian Grouping \cite{ye2025gaussian}. As illustrated in the provided visualization, OVGaussian demonstrates superior segmentation accuracy, particularly in handling complex scenes with overlapping objects and fine-grained details (\cref{fig:visual_3D_sup}). While CLIP2Scene often struggles with inconsistent boundaries and under-segmented regions, OVGaussian exhibits precise delineation of object boundaries and improved recognition of diverse semantic categories. Moreover, Gaussian Grouping, though effective in certain contexts, fails to generalize across unseen scenes, resulting in incomplete segmentations. 

\begin{figure*}
  %%%%\vspace*{-5ex}
  \centerline{\includegraphics[width=1\textwidth]{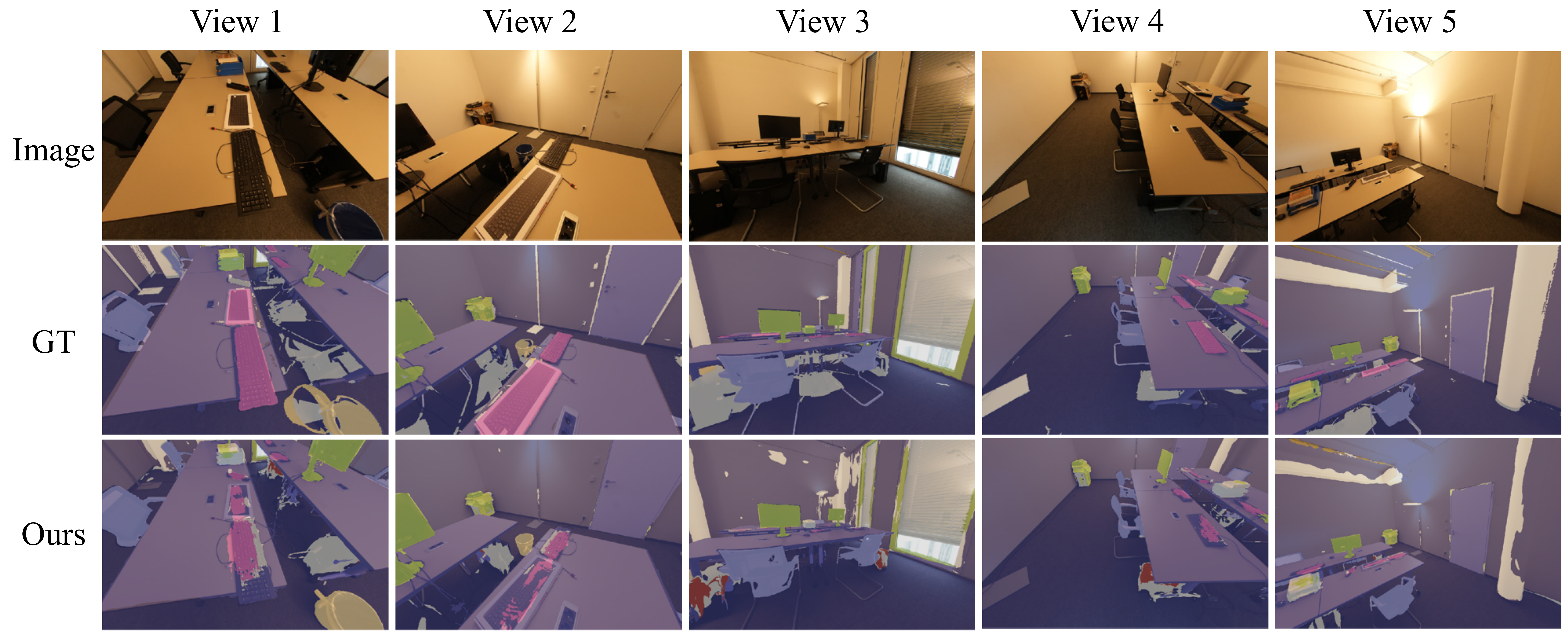}}
  %\vspace*{-0.5ex}
  \vspace{-2ex}
  \caption{Visualization of cross-view consistency in novel viewpoints. It illustrates OVGaussian’s ability to maintain coherent segmentation across diverse viewpoints, showcasing cross-view consistency in 3D scene understanding.}
  \vspace{-3ex}
  \label{fig:visual_2D_sup_1}
\end{figure*}

\begin{figure*}
  %%%%\vspace*{-5ex}
  \centerline{\includegraphics[width=1\textwidth]{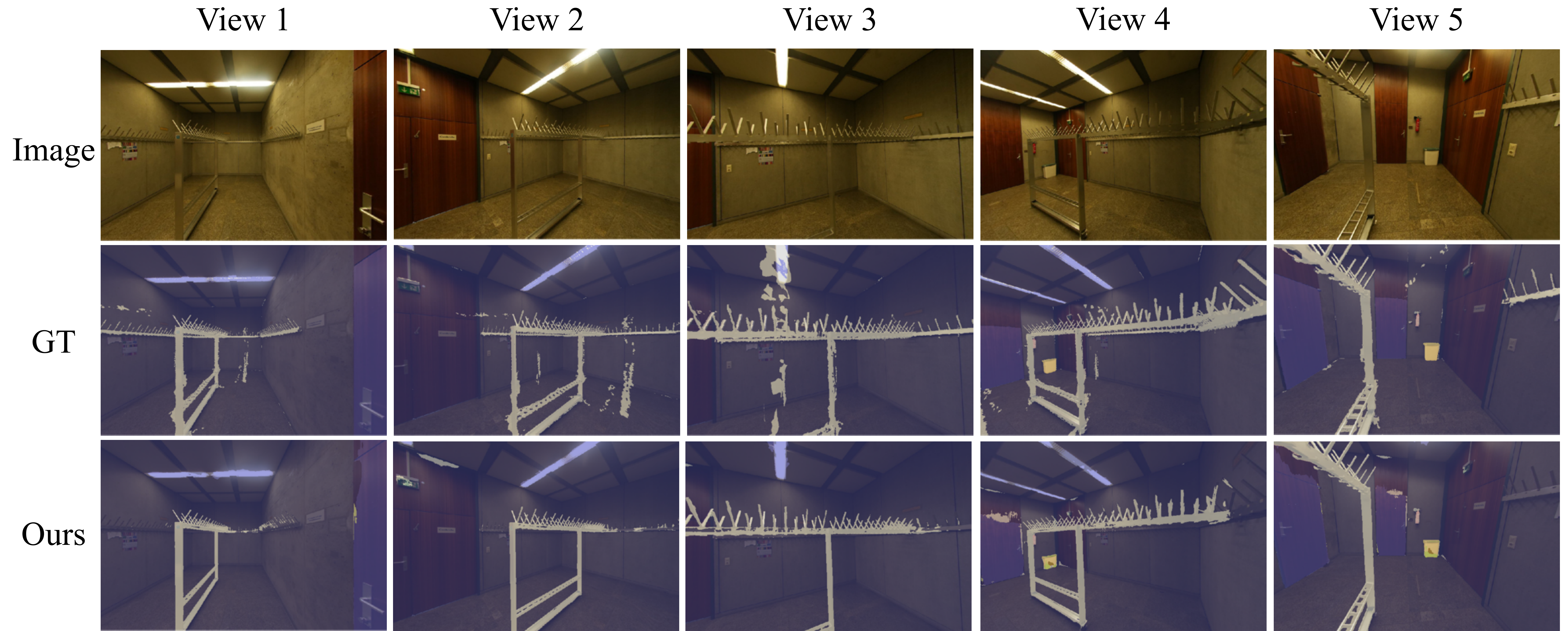}}
  %\vspace*{-0.5ex}
  \vspace{-2ex}
  \caption{Visualization of cross-view consistency in novel viewpoints. It illustrates OVGaussian’s ability to maintain coherent segmentation across diverse viewpoints, showcasing cross-view consistency in 3D scene understanding.}
  \vspace{-3ex}
  \label{fig:visual_2D_sup_2}
\end{figure*}

\begin{figure*}
  %%%%\vspace*{-5ex}
  \centerline{\includegraphics[width=1\textwidth]{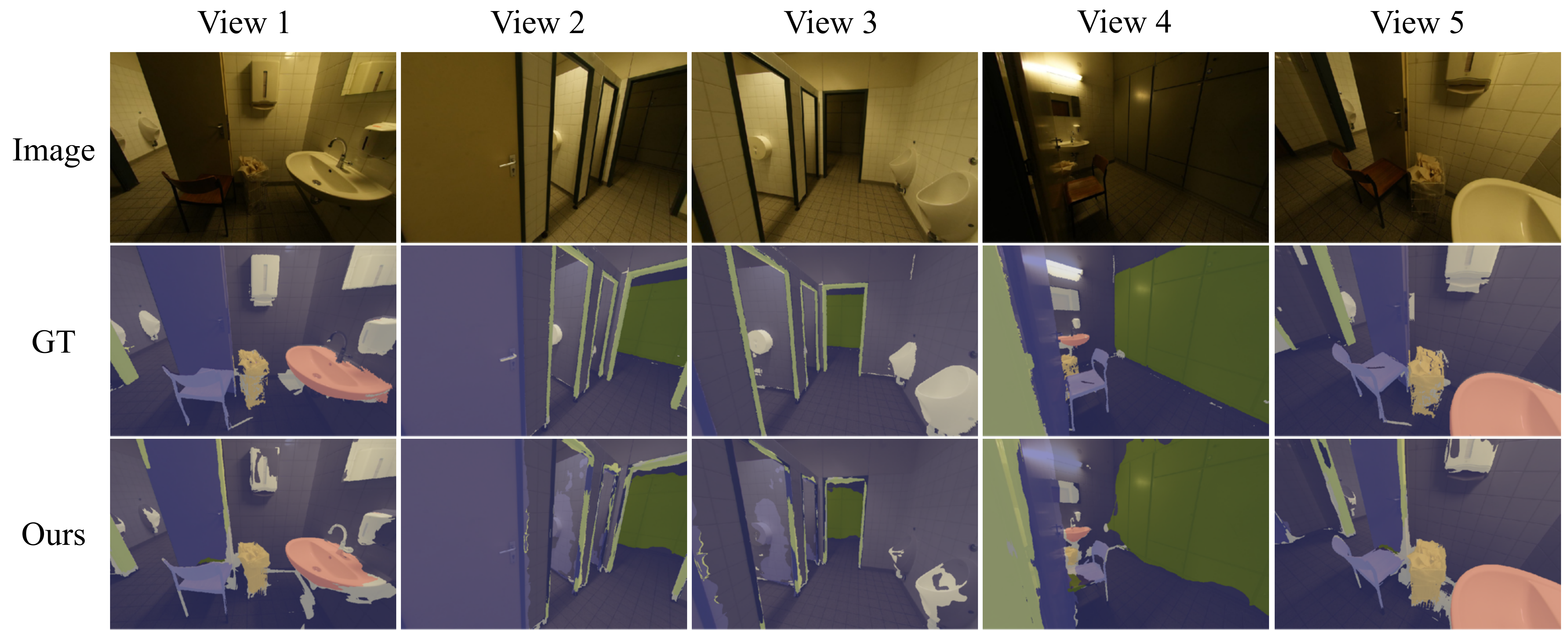}}
  %\vspace*{-0.5ex}
  \vspace{-2ex}
  \caption{Visualization of cross-view consistency in novel viewpoints. It illustrates OVGaussian’s ability to maintain coherent segmentation across diverse viewpoints, showcasing cross-view consistency in 3D scene understanding.}
  \vspace{-3ex}
  \label{fig:visual_2D_sup_3}
\end{figure*}

\begin{figure*}
  %%%%\vspace*{-5ex}
  \centerline{\includegraphics[width=1\textwidth]{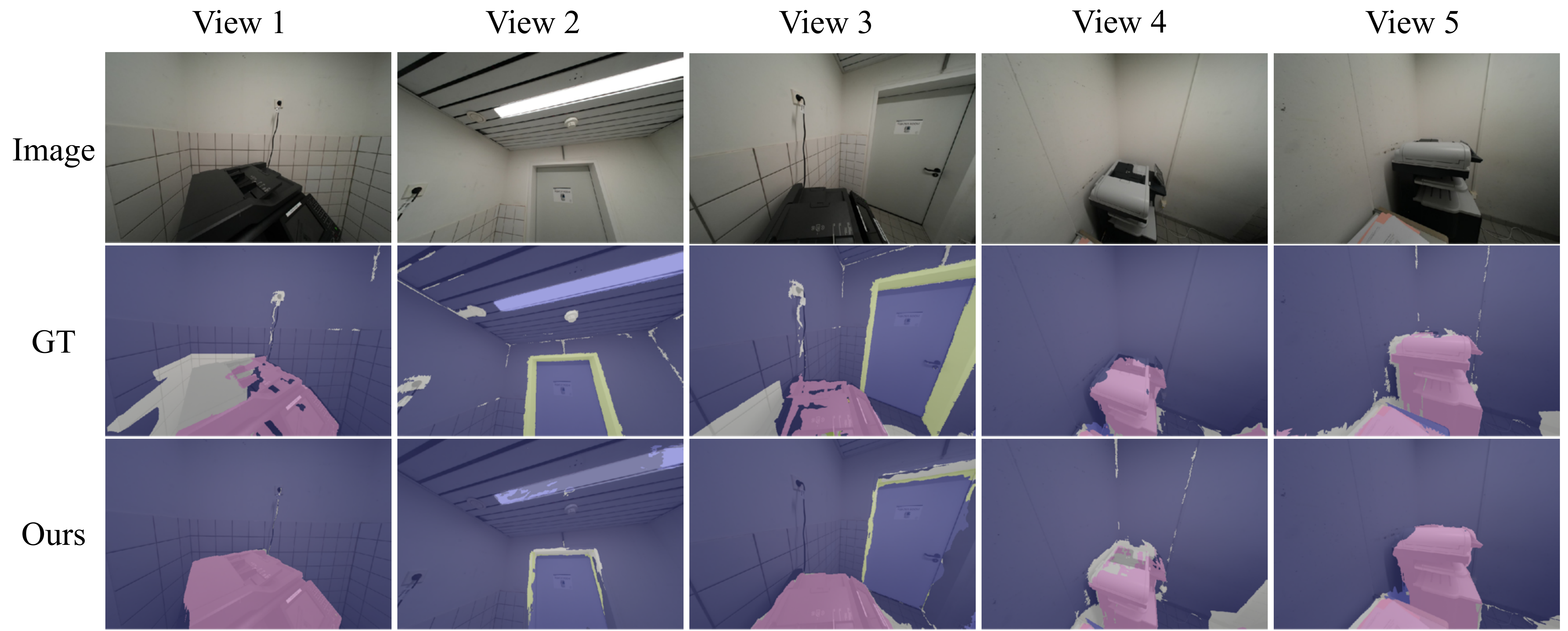}}
  %\vspace*{-0.5ex}
  \vspace{-2ex}
  \caption{Visualization of cross-view consistency in novel viewpoints. It illustrates OVGaussian’s ability to maintain coherent segmentation across diverse viewpoints, showcasing cross-view consistency in 3D scene understanding.}
  \vspace{-3ex}
  \label{fig:visual_2D_sup_4}
\end{figure*}

\begin{figure*}
  %%%%\vspace*{-5ex}
  \centerline{\includegraphics[width=1\textwidth]{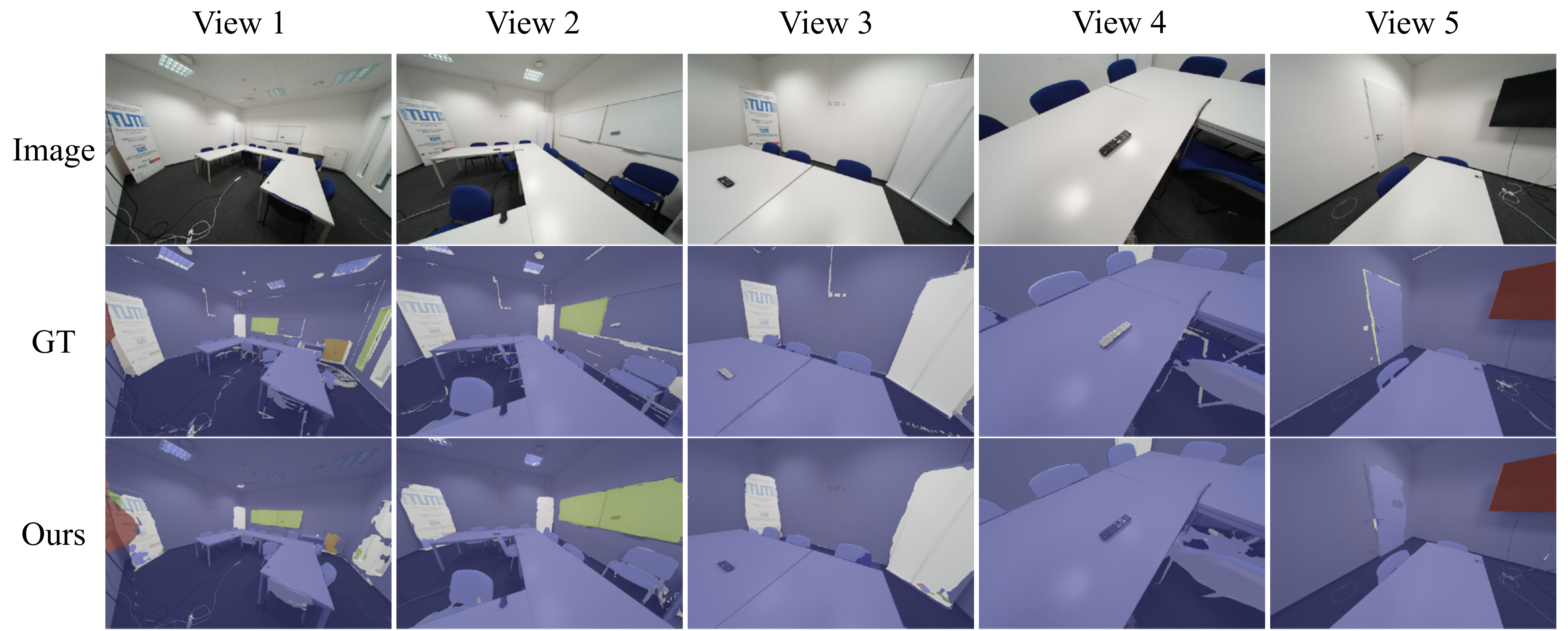}}
  %\vspace*{-0.5ex}
  \vspace{-2ex}
  \caption{Visualization of cross-view consistency in novel viewpoints. It illustrates OVGaussian’s ability to maintain coherent segmentation across diverse viewpoints, showcasing cross-view consistency in 3D scene understanding.}
  \vspace{-3ex}
  \label{fig:visual_2D_sup_5}
\end{figure*}

To evaluate the effectiveness of our method, we conducted a qualitative analysis across multiple views (View 1 to View 5), comparing the ground truth (GT) against our segmentation results (Fig. \ref{fig:visual_2D_sup_1}, \ref{fig:visual_2D_sup_2}, \ref{fig:visual_2D_sup_3}, \ref{fig:visual_2D_sup_4} and \ref{fig:visual_2D_sup_5}). As shown in the visualization, our method demonstrates robust multi-view consistency and accurate segmentation of diverse object categories. From the provided images, it is evident that our approach successfully captures fine-grained semantic details and aligns them across different viewpoints. While the input images exhibit significant variations in scene composition and object appearance, our method consistently maintains semantic coherence, producing precise and visually appealing segmentation results. Notably, our approach shows superior performance in handling complex object boundaries and maintaining category-specific segmentation consistency. This underscores the effectiveness of the proposed cross-modal alignment and the Gaussian-based representation in achieving high-quality, open-vocabulary segmentation across multiple perspectives.
Additionally, we provide a video demo, \textbf{“OVGaussian\_demo.mp4,”} as a supplementary material, which further illustrates the performance of our method across various scenes and viewpoints.

\section{Broader Impact \& Limitations}
\label{sec:limitations}
In this section, we discuss the broader impact and limitations of our work.

\subsection{Broader Impact}
The development of OVGaussian has the potential to advance open-vocabulary 3D scene understanding, enabling models to generalize across diverse environments, object categories, and viewpoints. This capability holds significant promise for a range of applications, including robotics, autonomous vehicles, augmented reality, and smart cities, where understanding complex and dynamic 3D scenes is crucial. However, the broader adoption of OVGaussian also raises certain considerations. The use of 3D scene understanding models in real-world applications, such as surveillance or autonomous systems, necessitates ethical considerations regarding privacy, safety, and accountability. Developers and stakeholders must ensure that these technologies are deployed responsibly and transparently, with safeguards in place to minimize misuse and unintended consequences. Overall, OVGaussian represents a step forward in bridging the gap between open-vocabulary understanding and scalable 3D scene analysis, fostering innovation while highlighting the importance of addressing ethical and societal implications in AI research.

\subsection{Potential Limitations}
Despite the strong performance demonstrated by our method, two key limitations remain:
\begin{itemize}
    \item \textbf{Dependency on 2D Supervision:} Our approach relies on 2D vision foundation models for knowledge transfer, which can limit performance when 2D annotations or pre-trained models are unavailable or poorly aligned with the target 3D domain. Future work could explore self-supervised or weakly-supervised techniques to reduce this dependency, enabling more robust and adaptable 3D scene understanding.

    \item \textbf{Scalability to Large-Scale Scenes:} Although the proposed Gaussian-based representation is computationally efficient, scaling to very large or open-world scenes remains challenging due to memory and processing constraints. Future enhancements could involve optimized hierarchical Gaussian representations or efficient point sampling techniques to handle large-scale datasets without sacrificing segmentation accuracy.
\end{itemize}

\section{Public Resource Used}
In this section, we acknowledge the use of the following public resources, during the course of this work:
\begin{itemize}
    \item ScanNet++\footnote{\url{https://kaldir.vc.in.tum.de/scannetpp}.}\cite{yeshwanth2023scannet++} \dotfill ScanNet++ License

    \item Replica\footnote{\url{https://github.com/facebookresearch/Replica-Dataset}.}\cite{straub2019replica} \dotfill Replica Dataset License

    \item CLIP2Scene\footnote{\url{https://github.com/runnanchen/CLIP2Scene}.} \cite{chen2023clip2scene} \dotfill Apache License 2.0
    
    \item Gaussian Grouping\footnote{\url{https://github.com/lkeab/gaussian-grouping}.} \cite{ye2025gaussian} \dotfill Apache License 2.0
    
    \item MaskCLIP\footnote{\url{https://github.com/chongzhou96/MaskCLIP}.} \cite{maskclip} \dotfill Apache License 2.0

    \item CLIP\footnote{\url{https://github.com/openai/CLIP}.} \cite{radford2021learning} \dotfill MIT License
    
\end{itemize}

\clearpage
\clearpage
{
    \small
    \bibliographystyle{ieeenat_fullname}
    \bibliography{main}

\begin{thebibliography}{59}
\providecommand{\natexlab}[1]{#1}
\providecommand{\url}[1]{\texttt{#1}}
\expandafter\ifx\csname urlstyle\endcsname\relax
  \providecommand{\doi}[1]{doi: #1}\else
  \providecommand{\doi}{doi: \begingroup \urlstyle{rm}\Url}\fi

\bibitem[Bucher et~al.(2019)Bucher, Vu, Cord, and P{\'e}rez]{bucher2019zero}
Maxime Bucher, Tuan-Hung Vu, Matthieu Cord, and Patrick P{\'e}rez.
\newblock Zero-shot semantic segmentation.
\newblock \emph{Advances in Neural Information Processing Systems}, 32, 2019.

\bibitem[Chen et~al.(2024{\natexlab{a}})Chen, Chen, Qu, Wang, Liu, Chen, and Chung]{chen2024beyond}
Haodong Chen, Runnan Chen, Qiang Qu, Zhaoqing Wang, Tongliang Liu, Xiaoming Chen, and Yuk~Ying Chung.
\newblock Beyond gaussians: Fast and high-fidelity 3d splatting with linear kernels.
\newblock \emph{arXiv preprint arXiv:2411.12440}, 2024{\natexlab{a}}.

\bibitem[Chen et~al.(2022)Chen, Zhu, Chen, Li, Ma, Yang, and Wang]{chen2022zero}
Runnan Chen, Xinge Zhu, Nenglun Chen, Wei Li, Yuexin Ma, Ruigang Yang, and Wenping Wang.
\newblock Zero-shot point cloud segmentation by transferring geometric primitives.
\newblock \emph{arXiv preprint arXiv:2210.09923}, 2022.

\bibitem[Chen et~al.(2023{\natexlab{a}})Chen, Liu, Kong, Zhu, Ma, Li, Hou, Qiao, and Wang]{chen2023clip2scene}
Runnan Chen, Youquan Liu, Lingdong Kong, Xinge Zhu, Yuexin Ma, Yikang Li, Yuenan Hou, Yu Qiao, and Wenping Wang.
\newblock Clip2scene: Towards label-efficient 3d scene understanding by clip.
\newblock In \emph{Proceedings of the IEEE/CVF Conference on Computer Vision and Pattern Recognition}, pages 7020--7030, 2023{\natexlab{a}}.

\bibitem[Chen et~al.(2023{\natexlab{b}})Chen, Zhu, Chen, Li, Ma, Yang, and Wang]{chen2023bridging}
Runnan Chen, Xinge Zhu, Nenglun Chen, Wei Li, Yuexin Ma, Ruigang Yang, and Wenping Wang.
\newblock Bridging language and geometric primitives for zero-shot point cloud segmentation.
\newblock In \emph{Proceedings of the 31st ACM International Conference on Multimedia}, pages 5380--5388, 2023{\natexlab{b}}.

\bibitem[Chen et~al.(2023{\natexlab{c}})Chen, Zhu, Chen, Wang, Li, Ma, Yang, Liu, and Wang]{chen2023model2scene}
Runnan Chen, Xinge Zhu, Nenglun Chen, Dawei Wang, Wei Li, Yuexin Ma, Ruigang Yang, Tongliang Liu, and Wenping Wang.
\newblock Model2scene: Learning 3d scene representation via contrastive language-cad models pre-training.
\newblock \emph{arXiv preprint arXiv:2309.16956}, 2023{\natexlab{c}}.

\bibitem[Chen et~al.(2024{\natexlab{b}})Chen, Liu, Kong, Chen, Zhu, Ma, Liu, and Wang]{chen2024towards}
Runnan Chen, Youquan Liu, Lingdong Kong, Nenglun Chen, Xinge Zhu, Yuexin Ma, Tongliang Liu, and Wenping Wang.
\newblock Towards label-free scene understanding by vision foundation models.
\newblock \emph{Advances in Neural Information Processing Systems}, 36, 2024{\natexlab{b}}.

\bibitem[Chen et~al.(2024{\natexlab{c}})Chen, Wang, Wang, and Liu]{chen2024text}
Zilong Chen, Feng Wang, Yikai Wang, and Huaping Liu.
\newblock Text-to-3d using gaussian splatting.
\newblock In \emph{Proceedings of the IEEE/CVF Conference on Computer Vision and Pattern Recognition}, pages 21401--21412, 2024{\natexlab{c}}.

\bibitem[Cheng et~al.(2020)Cheng, Collins, Zhu, Liu, Huang, Adam, and Chen]{cheng2020panoptic}
Bowen Cheng, Maxwell~D Collins, Yukun Zhu, Ting Liu, Thomas~S Huang, Hartwig Adam, and Liang-Chieh Chen.
\newblock Panoptic-deeplab: A simple, strong, and fast baseline for bottom-up panoptic segmentation.
\newblock In \emph{Proceedings of the IEEE/CVF conference on computer vision and pattern recognition}, pages 12475--12485, 2020.

\bibitem[Cheng et~al.(2021{\natexlab{a}})Cheng, Schwing, and Kirillov]{cheng2021per}
Bowen Cheng, Alex Schwing, and Alexander Kirillov.
\newblock Per-pixel classification is not all you need for semantic segmentation.
\newblock \emph{Advances in neural information processing systems}, 34:\penalty0 17864--17875, 2021{\natexlab{a}}.

\bibitem[Cheng et~al.(2022)Cheng, Misra, Schwing, Kirillov, and Girdhar]{cheng2022masked}
Bowen Cheng, Ishan Misra, Alexander~G Schwing, Alexander Kirillov, and Rohit Girdhar.
\newblock Masked-attention mask transformer for universal image segmentation.
\newblock In \emph{Proceedings of the IEEE/CVF Conference on Computer Vision and Pattern Recognition}, pages 1290--1299, 2022.

\bibitem[Cheng et~al.(2021{\natexlab{b}})Cheng, Razani, Taghavi, Li, and Liu]{af2s3net}
Ran Cheng, Ryan Razani, Ehsan Taghavi, Enxu Li, and Bingbing Liu.
\newblock (af)2-s3net: Attentive feature fusion with adaptive feature selection for sparse semantic segmentation network.
\newblock In \emph{IEEE Conference on Computer Vision and Pattern Recognition}, pages 12547--12556, 2021{\natexlab{b}}.

\bibitem[Choy et~al.(2019)Choy, Gwak, and Savarese]{choy20194d}
Christopher Choy, JunYoung Gwak, and Silvio Savarese.
\newblock 4d spatio-temporal convnets: Minkowski convolutional neural networks.
\newblock In \emph{IEEE/CVF Conference on Computer Vision and Pattern Recognition}, pages 3075--3084, 2019.

\bibitem[Contributors(2020)]{contributors2020mmdetection3d}
MMDetection3D Contributors.
\newblock Mmdetection3d: Openmmlab next-generation platform for general 3d object detection, 2020.

\bibitem[Ding et~al.(2022)Ding, Yang, Xue, Zhang, Bai, and Qi]{ding2022language}
Runyu Ding, Jihan Yang, Chuhui Xue, Wenqing Zhang, Song Bai, and Xiaojuan Qi.
\newblock Language-driven open-vocabulary 3d scene understanding.
\newblock \emph{arXiv preprint arXiv:2211.16312}, 2022.

\bibitem[Hong et~al.(2022)Hong, Kong, Zhou, Zhu, Li, and Liu]{hong2022dsnet}
Fangzhou Hong, Lingdong Kong, Hui Zhou, Xinge Zhu, Hongsheng Li, and Ziwei Liu.
\newblock Unified 3d and 4d panoptic segmentation via dynamic shifting network.
\newblock \emph{arXiv preprint arXiv:2203.07186}, 2022.

\bibitem[Hu et~al.(2020)Hu, Sclaroff, and Saenko]{hu2020uncertainty}
Ping Hu, Stan Sclaroff, and Kate Saenko.
\newblock Uncertainty-aware learning for zero-shot semantic segmentation.
\newblock \emph{Advances in Neural Information Processing Systems}, 33:\penalty0 21713--21724, 2020.

\bibitem[Kerbl et~al.(2023)Kerbl, Kopanas, Leimk{\"u}hler, and Drettakis]{kerbl20233d}
Bernhard Kerbl, Georgios Kopanas, Thomas Leimk{\"u}hler, and George Drettakis.
\newblock 3d gaussian splatting for real-time radiance field rendering.
\newblock \emph{ACM Transactions on Graphics}, 42\penalty0 (4):\penalty0 1--14, 2023.

\bibitem[Kirillov et~al.(2023)Kirillov, Mintun, Ravi, Mao, Rolland, Gustafson, Xiao, Whitehead, Berg, Lo, et~al.]{kirillov2023segment}
Alexander Kirillov, Eric Mintun, Nikhila Ravi, Hanzi Mao, Chloe Rolland, Laura Gustafson, Tete Xiao, Spencer Whitehead, Alexander~C Berg, Wan-Yen Lo, et~al.
\newblock Segment anything.
\newblock In \emph{Proceedings of the IEEE/CVF International Conference on Computer Vision}, pages 4015--4026, 2023.

\bibitem[Kong et~al.(2023{\natexlab{a}})Kong, Liu, Chen, Ma, Zhu, Li, Hou, Qiao, and Liu]{kong2023rethinking}
Lingdong Kong, Youquan Liu, Runnan Chen, Yuexin Ma, Xinge Zhu, Yikang Li, Yuenan Hou, Yu Qiao, and Ziwei Liu.
\newblock Rethinking range view representation for lidar segmentation.
\newblock In \emph{Proceedings of the IEEE/CVF International Conference on Computer Vision}, pages 228--240, 2023{\natexlab{a}}.

\bibitem[Kong et~al.(2023{\natexlab{b}})Kong, Liu, Li, Chen, Zhang, Ren, Pan, Chen, and Liu]{kong2023robo3d}
Lingdong Kong, Youquan Liu, Xin Li, Runnan Chen, Wenwei Zhang, Jiawei Ren, Liang Pan, Kai Chen, and Ziwei Liu.
\newblock Robo3d: Towards robust and reliable 3d perception against corruptions.
\newblock In \emph{Proceedings of the IEEE/CVF International Conference on Computer Vision}, pages 19994--20006, 2023{\natexlab{b}}.

\bibitem[Li et~al.(2022)Li, Weinberger, Belongie, Koltun, and Ranftl]{lilanguage}
Boyi Li, Kilian~Q Weinberger, Serge Belongie, Vladlen Koltun, and Rene Ranftl.
\newblock Language-driven semantic segmentation.
\newblock In \emph{International Conference on Learning Representations}, 2022.

\bibitem[Li et~al.(2020)Li, Wei, and Yang]{li2020consistent}
Peike Li, Yunchao Wei, and Yi Yang.
\newblock Consistent structural relation learning for zero-shot segmentation.
\newblock \emph{Advances in Neural Information Processing Systems}, 33:\penalty0 10317--10327, 2020.

\bibitem[Liu et~al.(2023)Liu, Chen, Li, Kong, Yang, Xia, Bai, Zhu, Ma, Li, et~al.]{liu2023uniseg}
Youquan Liu, Runnan Chen, Xin Li, Lingdong Kong, Yuchen Yang, Zhaoyang Xia, Yeqi Bai, Xinge Zhu, Yuexin Ma, Yikang Li, et~al.
\newblock Uniseg: A unified multi-modal lidar segmentation network and the openpcseg codebase.
\newblock In \emph{Proceedings of the IEEE/CVF International Conference on Computer Vision}, pages 21662--21673, 2023.

\bibitem[Liu et~al.(2024)Liu, Kong, Wu, Chen, Li, Pan, Liu, and Ma]{liu2024multi}
Youquan Liu, Lingdong Kong, Xiaoyang Wu, Runnan Chen, Xin Li, Liang Pan, Ziwei Liu, and Yuexin Ma.
\newblock Multi-space alignments towards universal lidar segmentation.
\newblock In \emph{Proceedings of the IEEE/CVF Conference on Computer Vision and Pattern Recognition}, pages 14648--14661, 2024.

\bibitem[Lu et~al.(2023)Lu, Jiang, Chen, Hou, Zhu, and Ma]{lu2023see}
Yuhang Lu, Qi Jiang, Runnan Chen, Yuenan Hou, Xinge Zhu, and Yuexin Ma.
\newblock See more and know more: Zero-shot point cloud segmentation via multi-modal visual data.
\newblock In \emph{Proceedings of the IEEE/CVF International Conference on Computer Vision}, pages 21674--21684, 2023.

\bibitem[Luiten et~al.(2024)Luiten, Kopanas, Leibe, and Ramanan]{luiten2024dynamic}
Jonathon Luiten, Georgios Kopanas, Bastian Leibe, and Deva Ramanan.
\newblock Dynamic 3d gaussians: Tracking by persistent dynamic view synthesis.
\newblock In \emph{2024 International Conference on 3D Vision (3DV)}, pages 800--809. IEEE, 2024.

\bibitem[Michele et~al.(2021)Michele, Boulch, Puy, Bucher, and Marlet]{michele2021generative}
Bj{\"o}rn Michele, Alexandre Boulch, Gilles Puy, Maxime Bucher, and Renaud Marlet.
\newblock Generative zero-shot learning for semantic segmentation of 3d point clouds.
\newblock In \emph{International Conference on 3D Vision}, pages 992--1002, 2021.

\bibitem[Peng et~al.(2023)Peng, Genova, Jiang, Tagliasacchi, Pollefeys, Funkhouser, et~al.]{peng2023openscene}
Songyou Peng, Kyle Genova, Chiyu Jiang, Andrea Tagliasacchi, Marc Pollefeys, Thomas Funkhouser, et~al.
\newblock Openscene: 3d scene understanding with open vocabularies.
\newblock In \emph{Proceedings of the IEEE/CVF conference on computer vision and pattern recognition}, pages 815--824, 2023.

\bibitem[Peng et~al.(2025)Peng, Chen, Qiao, Kong, Liu, Sun, Wang, Zhu, and Ma]{peng2025learning}
Xidong Peng, Runnan Chen, Feng Qiao, Lingdong Kong, Youquan Liu, Yujing Sun, Tai Wang, Xinge Zhu, and Yuexin Ma.
\newblock Learning to adapt sam for segmenting cross-domain point clouds.
\newblock In \emph{European Conference on Computer Vision}, pages 54--71. Springer, 2025.

\bibitem[Qi et~al.(2017)Qi, Su, Mo, and Guibas]{qi2017pointnet}
Charles~R Qi, Hao Su, Kaichun Mo, and Leonidas~J Guibas.
\newblock Pointnet: Deep learning on point sets for 3d classification and segmentation.
\newblock In \emph{IEEE/CVF Conference on Computer Vision and Pattern Recognition}, pages 652--660, 2017.

\bibitem[Qin et~al.(2024)Qin, Li, Zhou, Wang, and Pfister]{qin2024langsplat}
Minghan Qin, Wanhua Li, Jiawei Zhou, Haoqian Wang, and Hanspeter Pfister.
\newblock Langsplat: 3d language gaussian splatting.
\newblock In \emph{Proceedings of the IEEE/CVF Conference on Computer Vision and Pattern Recognition}, pages 20051--20060, 2024.

\bibitem[Radford et~al.(2021)Radford, Kim, Hallacy, Ramesh, Goh, Agarwal, Sastry, Askell, Mishkin, Clark, et~al.]{radford2021learning}
Alec Radford, Jong~Wook Kim, Chris Hallacy, Aditya Ramesh, Gabriel Goh, Sandhini Agarwal, Girish Sastry, Amanda Askell, Pamela Mishkin, Jack Clark, et~al.
\newblock Learning transferable visual models from natural language supervision.
\newblock In \emph{International Conference on Machine Learning}, pages 8748--8763. PMLR, 2021.

\bibitem[Riz et~al.(2023)Riz, Saltori, Ricci, and Poiesi]{riz2023novel}
Luigi Riz, Cristiano Saltori, Elisa Ricci, and Fabio Poiesi.
\newblock Novel class discovery for 3d point cloud semantic segmentation.
\newblock \emph{arXiv preprint arXiv:2303.11610}, 2023.

\bibitem[Sautier et~al.(2022)Sautier, Puy, Gidaris, Boulch, Bursuc, and Marlet]{sautier2022image}
Corentin Sautier, Gilles Puy, Spyros Gidaris, Alexandre Boulch, Andrei Bursuc, and Renaud Marlet.
\newblock Image-to-lidar self-supervised distillation for autonomous driving data.
\newblock In \emph{Proceedings of the IEEE/CVF Conference on Computer Vision and Pattern Recognition}, pages 9891--9901, 2022.

\bibitem[Shi et~al.(2024)Shi, Wang, Duan, and Guan]{shi2024language}
Jin-Chuan Shi, Miao Wang, Hao-Bin Duan, and Shao-Hua Guan.
\newblock Language embedded 3d gaussians for open-vocabulary scene understanding.
\newblock In \emph{Proceedings of the IEEE/CVF Conference on Computer Vision and Pattern Recognition}, pages 5333--5343, 2024.

\bibitem[Straub et~al.(2019)Straub, Whelan, Ma, Chen, Wijmans, Green, Engel, Mur-Artal, Ren, Verma, et~al.]{straub2019replica}
Julian Straub, Thomas Whelan, Lingni Ma, Yufan Chen, Erik Wijmans, Simon Green, Jakob~J Engel, Raul Mur-Artal, Carl Ren, Shobhit Verma, et~al.
\newblock The replica dataset: A digital replica of indoor spaces.
\newblock \emph{arXiv preprint arXiv:1906.05797}, 2019.

\bibitem[Strudel et~al.(2021)Strudel, Garcia, Laptev, and Schmid]{strudel2021segmenter}
Robin Strudel, Ricardo Garcia, Ivan Laptev, and Cordelia Schmid.
\newblock Segmenter: Transformer for semantic segmentation.
\newblock In \emph{Proceedings of the IEEE/CVF international conference on computer vision}, pages 7262--7272, 2021.

\bibitem[Sun et~al.(2024)Sun, Qing, Xu, Kong, Liu, Li, Zhu, Zhang, Xiao, Chen, et~al.]{sun2024empirical}
Jiahao Sun, Chunmei Qing, Xiang Xu, Lingdong Kong, Youquan Liu, Li Li, Chenming Zhu, Jingwei Zhang, Zeqi Xiao, Runnan Chen, et~al.
\newblock An empirical study of training state-of-the-art lidar segmentation models.
\newblock \emph{arXiv preprint arXiv:2405.14870}, 2024.

\bibitem[Tang et~al.(2023)Tang, Ren, Zhou, Liu, and Zeng]{tang2023dreamgaussian}
Jiaxiang Tang, Jiawei Ren, Hang Zhou, Ziwei Liu, and Gang Zeng.
\newblock Dreamgaussian: Generative gaussian splatting for efficient 3d content creation.
\newblock \emph{arXiv preprint arXiv:2309.16653}, 2023.

\bibitem[Vaswani et~al.(2017)Vaswani, Shazeer, Parmar, Uszkoreit, Jones, Gomez, Kaiser, and Polosukhin]{vaswani2017attention}
Ashish Vaswani, Noam Shazeer, Niki Parmar, Jakob Uszkoreit, Llion Jones, Aidan~N Gomez, {\L}ukasz Kaiser, and Illia Polosukhin.
\newblock Attention is all you need.
\newblock In \emph{Advances in Neural Information Processing Systems}, 2017.

\bibitem[Wu et~al.(2024)Wu, Yi, Fang, Xie, Zhang, Wei, Liu, Tian, and Wang]{wu20244d}
Guanjun Wu, Taoran Yi, Jiemin Fang, Lingxi Xie, Xiaopeng Zhang, Wei Wei, Wenyu Liu, Qi Tian, and Xinggang Wang.
\newblock 4d gaussian splatting for real-time dynamic scene rendering.
\newblock In \emph{Proceedings of the IEEE/CVF Conference on Computer Vision and Pattern Recognition}, pages 20310--20320, 2024.

\bibitem[Wu et~al.(2022)Wu, Lao, Jiang, Liu, and Zhao]{wu2022point}
Xiaoyang Wu, Yixing Lao, Li Jiang, Xihui Liu, and Hengshuang Zhao.
\newblock Point transformer v2: Grouped vector attention and partition-based pooling.
\newblock \emph{arXiv preprint arXiv:2210.05666}, 2022.

\bibitem[Xu et~al.(2021{\natexlab{a}})Xu, Zhang, Dou, Zhu, Sun, and Pu]{rpvnet}
Jianyun Xu, Ruixiang Zhang, Jian Dou, Yushi Zhu, Jie Sun, and Shiliang Pu.
\newblock Rpvnet: A deep and efficient range-point-voxel fusion network for lidar point cloud segmentation.
\newblock In \emph{IEEE/CVF International Conference on Computer Vision}, pages 16024--16033, 2021{\natexlab{a}}.

\bibitem[Xu et~al.(2021{\natexlab{b}})Xu, Zhang, Wei, Lin, Cao, Hu, and Bai]{xu2021simple}
Mengde Xu, Zheng Zhang, Fangyun Wei, Yutong Lin, Yue Cao, Han Hu, and Xiang Bai.
\newblock A simple baseline for zero-shot semantic segmentation with pre-trained vision-language model.
\newblock \emph{arXiv preprint arXiv:2112.14757}, 2021{\natexlab{b}}.

\bibitem[Xu et~al.(2023)Xu, Cong, Yao, Chen, Hou, Zhu, He, Yu, and Ma]{xu2023human}
Yiteng Xu, Peishan Cong, Yichen Yao, Runnan Chen, Yuenan Hou, Xinge Zhu, Xuming He, Jingyi Yu, and Yuexin Ma.
\newblock Human-centric scene understanding for 3d large-scale scenarios.
\newblock In \emph{Proceedings of the IEEE/CVF International Conference on Computer Vision}, pages 20349--20359, 2023.

\bibitem[Yan et~al.(2022)Yan, Gao, Zheng, Zheng, Zhang, Cui, and Li]{XuYan20222DPASS2P}
Xu Yan, Jiantao Gao, Chaoda Zheng, Chaoda Zheng, Ruimao Zhang, Shenghui Cui, and Zhen Li.
\newblock 2dpass: 2d priors assisted semantic segmentation on lidar point clouds.
\newblock In \emph{ECCV}, 2022.

\bibitem[Yang et~al.(2023)Yang, Yang, Pan, and Zhang]{yang2023real}
Zeyu Yang, Hongye Yang, Zijie Pan, and Li Zhang.
\newblock Real-time photorealistic dynamic scene representation and rendering with 4d gaussian splatting.
\newblock \emph{arXiv preprint arXiv:2310.10642}, 2023.

\bibitem[Yang et~al.(2024)Yang, Gao, Zhou, Jiao, Zhang, and Jin]{yang2024deformable}
Ziyi Yang, Xinyu Gao, Wen Zhou, Shaohui Jiao, Yuqing Zhang, and Xiaogang Jin.
\newblock Deformable 3d gaussians for high-fidelity monocular dynamic scene reconstruction.
\newblock In \emph{Proceedings of the IEEE/CVF Conference on Computer Vision and Pattern Recognition}, pages 20331--20341, 2024.

\bibitem[Ye et~al.(2025)Ye, Danelljan, Yu, and Ke]{ye2025gaussian}
Mingqiao Ye, Martin Danelljan, Fisher Yu, and Lei Ke.
\newblock Gaussian grouping: Segment and edit anything in 3d scenes.
\newblock In \emph{European Conference on Computer Vision}, pages 162--179. Springer, 2025.

\bibitem[Yeshwanth et~al.(2023)Yeshwanth, Liu, Nie{\ss}ner, and Dai]{yeshwanth2023scannet++}
Chandan Yeshwanth, Yueh-Cheng Liu, Matthias Nie{\ss}ner, and Angela Dai.
\newblock Scannet++: A high-fidelity dataset of 3d indoor scenes.
\newblock In \emph{Proceedings of the IEEE/CVF International Conference on Computer Vision}, pages 12--22, 2023.

\bibitem[Yi et~al.(2023)Yi, Fang, Wu, Xie, Zhang, Liu, Tian, and Wang]{yi2023gaussiandreamer}
Taoran Yi, Jiemin Fang, Guanjun Wu, Lingxi Xie, Xiaopeng Zhang, Wenyu Liu, Qi Tian, and Xinggang Wang.
\newblock Gaussiandreamer: Fast generation from text to 3d gaussian splatting with point cloud priors.
\newblock \emph{arXiv preprint arXiv:2310.08529}, 2023.

\bibitem[Yin et~al.(2024)Yin, Shen, Chen, Li, Yang, Frossard, and Wang]{yin2024fusion}
Junbo Yin, Jianbing Shen, Runnan Chen, Wei Li, Ruigang Yang, Pascal Frossard, and Wenguan Wang.
\newblock Is-fusion: Instance-scene collaborative fusion for multimodal 3d object detection.
\newblock In \emph{Proceedings of the IEEE/CVF Conference on Computer Vision and Pattern Recognition}, pages 14905--14915, 2024.

\bibitem[Zhang and Ding(2021)]{zhang2021prototypical}
Hui Zhang and Henghui Ding.
\newblock Prototypical matching and open set rejection for zero-shot semantic segmentation.
\newblock In \emph{Proceedings of the IEEE/CVF International Conference on Computer Vision}, pages 6974--6983, 2021.

\bibitem[Zhou et~al.(2022{\natexlab{a}})Zhou, Loy, and Dai]{maskclip}
Chong Zhou, Chen~Change Loy, and Bo Dai.
\newblock Extract free dense labels from clip.
\newblock In \emph{European Conference on Computer Vision}, pages 696--712, 2022{\natexlab{a}}.

\bibitem[Zhou et~al.(2022{\natexlab{b}})Zhou, Loy, and Dai]{zhou2022maskclip}
Chong Zhou, Chen~Change Loy, and Bo Dai.
\newblock Extract free dense labels from clip.
\newblock In \emph{European Conference on Computer Vision (ECCV)}, 2022{\natexlab{b}}.

\bibitem[Zhou et~al.(2024)Zhou, Chang, Jiang, Fan, Zhu, Xu, Chari, You, Wang, and Kadambi]{zhou2024feature}
Shijie Zhou, Haoran Chang, Sicheng Jiang, Zhiwen Fan, Zehao Zhu, Dejia Xu, Pradyumna Chari, Suya You, Zhangyang Wang, and Achuta Kadambi.
\newblock Feature 3dgs: Supercharging 3d gaussian splatting to enable distilled feature fields.
\newblock In \emph{Proceedings of the IEEE/CVF Conference on Computer Vision and Pattern Recognition}, pages 21676--21685, 2024.

\bibitem[Zhu et~al.(2020)Zhu, Zhou, Wang, Hong, Ma, Li, Li, and Lin]{zhu2020cylindrical}
Xinge Zhu, Hui Zhou, Tai Wang, Fangzhou Hong, Yuexin Ma, Wei Li, Hongsheng Li, and Dahua Lin.
\newblock Cylindrical and asymmetrical 3d convolution networks for lidar segmentation.
\newblock \emph{arXiv preprint arXiv:2011.10033}, 2020.

\bibitem[Zhu et~al.(2021)Zhu, Zhou, Wang, Hong, Ma, Li, Li, and Lin]{zhu2021cylindrical}
Xinge Zhu, Hui Zhou, Tai Wang, Fangzhou Hong, Yuexin Ma, Wei Li, Hongsheng Li, and Dahua Lin.
\newblock Cylindrical and asymmetrical 3d convolution networks for lidar segmentation.
\newblock In \emph{IEEE/CVF Conference on Computer Vision and Pattern Recognition}, pages 9939--9948, 2021.

\end{thebibliography}
}


\end{document}